\documentclass[10pt,twocolumn,letterpaper]{article}

\usepackage[pagenumbers]{cvpr} 

\usepackage[dvipsnames]{xcolor}

\definecolor{cvprblue}{rgb}{0.21,0.49,0.74}
\usepackage[pagebackref,breaklinks,colorlinks,citecolor=cvprblue]{hyperref}

\def\paperID{1179} 
\def\confName{CVPR}
\def\confYear{2024}

\title{High-Quality Facial Geometry and Appearance Capture at Home}

\author{
Yuxuan Han
\and Junfeng Lyu
\and Feng Xu
\smallskip
\and
School of Software and BNRist, Tsinghua University
}

\begin{document}

\twocolumn[{%
\renewcommand\twocolumn[1][]{#1}%
\maketitle

\begin{center}
    \centering
    \captionsetup{type=figure}
    \includegraphics[width=\textwidth]{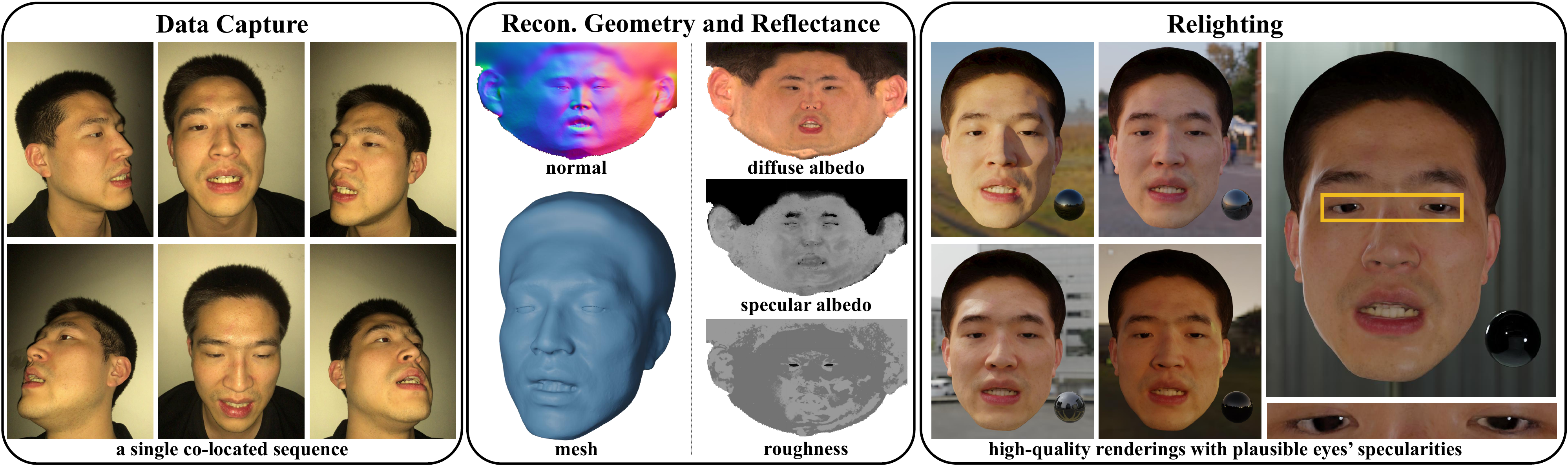}
    \vspace{-15pt}
    \captionof{figure}{
    We propose a novel method for high-quality face capture, featuring a low-cost and easy-to-use capture setup and the capability to model the complete face with skin, mouth interior, hair, and eyes. 
    Our method takes a single co-located smartphone flashlight sequence captured in a dim room (\emph{e.g.} rooms with curtains or at night) as input. It reconstructs relightable 3D face assets from the recorded data. These can be used by common graphics software like Blender to create photo-realistic renderings in new environments. }
    \label{Fig:teaser}
\end{center}%
}]

\begin{abstract}
\vspace{-11pt}
Facial geometry and appearance capture have demonstrated tremendous success in 3D scanning real humans in studios.
Recent works propose to democratize this technique while keeping the results high quality.
However, they are still inconvenient for daily usage. 
In addition, they focus on an easier problem of only capturing facial skin.
This paper proposes a novel method for high-quality face capture, featuring an easy-to-use system and the capability to model the complete face with skin, mouth interior, hair, and eyes.
We reconstruct facial geometry and appearance from a single co-located smartphone flashlight sequence captured in a dim room where the flashlight is the dominant light source (\emph{e.g.} rooms with curtains or at night).
To model the complete face, we propose a novel hybrid representation to effectively model both eyes and other facial regions, along with novel techniques to learn it from images.
We apply a combined lighting model to compactly represent real illuminations and exploit a morphable face albedo model as a reflectance prior to disentangle diffuse and specular. 
Experiments show that our method can capture high-quality 3D relightable scans.
Our code will be released \href{https://github.com/yxuhan/CoRA}{\textcolor{magenta}{here}}.
\end{abstract}

\section{Introduction}
High-quality facial geometry and appearance capture are the core steps for cloning our human beings to get into the digital world.
To achieve this, existing works~\cite{debevec2000acquiring,ma2007rapid,ghosh2011multiview,weyrich2006analysis,riviere2020single} develop specialized and expensive apparatus in studios to 3D scan real humans.
Although impressive results are demonstrated~\cite{alexander2009digital,alexander2013digital}, these techniques are currently only viable for a small number of professional users as on-site data capture is inconvenient and costly.
Thus, low-cost but high-quality face capture is strongly in demand to connect broad daily users to the digital world.

A few recent works~\cite{wang2023sunstage,azinovic2023high} focus on democratizing the face capture process while keeping the results as close as possible to the studio-based techniques.
The rationale of these methods is to exploit the high-frequency light sources in daily life, which is the key to recovering high-quality reflectance~\cite{ramamoorthi2001signal}.
SunStage~\cite{wang2023sunstage} exploits the sunlight where they reconstruct facial geometry and reflectance from a single selfie video of the subject rotating under the sun.
PolFace~\cite{azinovic2023high} exploits the smartphone flashlight, \emph{i.e.} they capture two co-located smartphone flashlight sequences with different polarization orientations in a darkroom to estimate facial geometry and reflectance.
Although both methods ease the face capture process to a large extent compared to the studio, the requirements of sunlight~\cite{wang2023sunstage}, polarization filter~\cite{azinovic2023high}, or darkroom~\cite{azinovic2023high} are still inconvenient for daily users to capture faces at home. 
In addition, as the reflectance property varies significantly across the face (\emph{e.g.} the almost-rough skin \emph{v.s.} the highly-specular eyes), these methods focus on an easier problem to only capture facial skin.

In this paper, we propose a novel method for low-cost high-quality facial geometry and appearance capture, which can model the complete face with skin, mouth interior, hair, and eyes.
Firstly, we propose a novel hybrid face representation to adopt different models for different facial regions, \emph{i.e.} eyeballs and other facial regions, due to their reflectance differences. 
For eyeballs,
we adopt two sphere meshes with predefined specular reflectance while leaving the spatially varying diffuse albedo to be solved from the recorded data.
The use of eyeball priors improves the reconstruction quality significantly since recovering geometry and reflectance for highly reflective objects (eyeballs in our case) is very challenging~\cite{verbin2022ref,liu2023nero}.
For other facial regions including skin, mouth interior, and hair, we adopt neural field~\cite{xie2022neural} considering its superior representation power and flexibility.
Specifically, we adopt a neural SDF field~\cite{yariv2021volume} to represent geometry and a neural field to model the parameters of the Disney BRDF~\cite{burley2012physically} as reflectance similar to previous works~\cite{zhang2022iron,cheng2023wildlight}.
To learn our hybrid representation from images, we design a novel mesh-aware volume rendering technique to integrate the eyeball meshes into the volume rendering process of the neural SDF field seamlessly.

To make our method easily used at home, we propose to train our model from a single co-located smartphone flashlight sequence captured in a dim room where the flashlight is the dominant light source (\emph{e.g.} rooms with curtains or at night).
Compared to previous works~\cite{azinovic2023high,wang2023sunstage}, our capture setup neither needs special equipment like the polarization filter and darkroom nor outdoor light sources like sunlight, making it more user-friendly.
However, it poses a new challenge to disentangle reflectance from the observed colors.
To this end, we involve both lighting and appearance priors to restrict the optimization. 
Firstly, we apply a combined lighting model to compactly represent both the low-frequency dim ambient light and the high-frequency smartphone flashlight.
Then, to constrain the diffuse-specular disentanglement, we resort to the reflectance prior provided by AlbedoMM~\cite{smith2020morphable}, a 3D morphable face albedo model trained on Light Stage scans~\cite{ghosh2011multiview,ma2007rapid,stratou2011effect}.
After training, we export our hybrid face representation to 3D assets compatible with common CG software (see Figure~\ref{Fig:teaser}).
By combining our method with Reflectance Transfer~\cite{peers2007post}, we demonstrate application on relightable facical performance capture in a low-cost setup.
Our main contributions include:
\begin{itemize}
    \item We propose a novel method for high-quality facial geometry and appearance capture, featuring a low-cost and easy-to-use capture setup and the capability to model the complete face with skin, mouth interior, hair, and eyes.
    \item We propose a novel hybrid representation to effectively model eyes and other facial regions and novel techniques to train it from images.
    \item We apply a combined lighting model to compactly represent the real illuminations and propose a reflectance constraint derived from AlbedoMM~\cite{smith2020morphable} to improve diffuse-specular disentanglement in our low-cost capture setup.
\end{itemize}

\section{Related Work}
\textbf{Face Capture.}
Face capture has attracted much attention in the past two decades.
Traditional methods have demonstrated very impressive results~\cite{alexander2009digital,alexander2013digital} under the studio-capture setup.
The seminal work of Debevec et al.~\cite{debevec2000acquiring} proposes to reconstruct the face reflectance fields by densely capturing One-Light-At-a-Time (OLAT) images of the human face using the Light Stage~\cite{debevec2012light}.
To capture 3D assets compatible with the graphics pipeline, Ma et al.~\cite{ma2007rapid} propose to capture the normal and albedo maps of faces leveraging polarized spherical gradient illumination.
Subsequently, Ghosh et al.~\cite{ghosh2011multiview} extend this technique to support multi-view capture to obtain ear-to-ear assets.
Another class of works captures faces under the single-shot setup.
Beeler et al.~\cite{beeler2010high} propose the first single-shot system to capture high-quality facial geometry, and then the follow-up works~\cite{gotardo2018practical,riviere2020single} extend it to support appearance capture.
However, all these methods require the users to travel to the studio for on-site capture, which is neither convenient nor low-cost for the broad daily users.

More recently, some works have proposed to democratize the process of face capture.
Some methods propose to reconstruct facial geometry and reflectance from a single in-the-wild image~\cite{lattas2020avatarme,yamaguchi2018high,Paraperas_2023_ICCV,han2023learning,smith2020morphable,li2020learning, lattas2023fitme,dib2021practical}.
Although these methods are easy to use for daily users, the reconstruction quality is far behind the studio-capture methods.
Another class of works attempts to capture faces in the multi-view setup.
NeuFace~\cite{zheng2023neuface} proposes to learn the facial geometry and a novel neural BRDF from the multi-view images captured under unknown low-frequency light.
To keep the face capture results as close as possible to the studio, recent methods propose to exploit high-frequency light sources in daily life like sunlight~\cite{wang2023sunstage} or smartphone flashlight~\cite{azinovic2023high}.
They solve facial geometry and reflectance from a single selfie video of the subject rotating under the sun~\cite{wang2023sunstage} or two co-located sequences with different polarization orientations captured in a darkroom~\cite{azinovic2023high}.
In this paper, we use only a single co-located sequence for face capture, which is more easy to use by daily users. 
We propose a novel hybrid representation by combining neural SDF field and mesh to reconstruct high-quality complete facial geometry and appearance including skin, mouth interior, hair, and eyes.

\textbf{Neural Fields for Inverse Rendering.}
Recent advances represent 3D scene attributes (\emph{e.g.} density and color) as a continuous function, \emph{a.k.a} neural fields~\cite{xie2022neural}, achieving state-of-the-art results on various tasks including view synthesis~\cite{muller2022instant,mildenhall2020nerf,barron2021mip,barron2022mip,barron2023zipnerf} and geometry reconstruction~\cite{wang2021neus,yariv2021volume,oechsle2021unisurf,yariv2020multiview}. 
More recently, some works~\cite{cheng2023wildlight,zhang2022iron,ling2023shadowneus,zhang2021nerfactor,zhang2021physg,bi2020neural,boss2021nerd,Munkberg_2022_CVPR,hasselgren2022shape} extend neural fields to inverse rendering, where geometry and reflectance are modeled as neural fields and learned from the captured images.
Among these works, the most relevant to us is WildLight~\cite{cheng2023wildlight}.
It solves geometry and reflectance from two sequences, one with the flashlight turned on and one turned off.
Similar to WildLight, we adopt a neural SDF field~\cite{yariv2021volume,wang2021neus} to represent geometry and a neural reflectance field to model the parameters of the Disney BRDF~\cite{burley2012physically}.
However, we apply a more compact lighting representation so that we require only a single flashlight turned-on sequence for training.
In addition, as we focus on the human face rather than common objects, we can exploit face priors.
We propose a hybrid representation to exploit eyeball priors to help reconstruction and a reflectance constraint derived from AlbedoMM~\cite{smith2020morphable} to regularize the neural reflectance fields.

\textbf{Hybrid Representation for Digital Avatar.}
Recent works~\cite{li2022eyenerf,zheng2023avatarrex,feng2022capturing} propose hybrid representation to model digital avatar, considering that we humans are made of components of different properties, \emph{e.g.} skin, hair, eyes, and clothes.
Among these works, the most relevant to us is EyeNeRF~\cite{li2022eyenerf}.
In EyeNeRF, the predefined eyeball meshes are used to guide the volume rendering process to better model the ray reflection and refraction on the eyeball surface; the facial geometry is totally represented by the neural density fields~\cite{mildenhall2020nerf}.
However, in our method, we adopt the eyeball mesh as part of the facial geometry and propose novel strategies to constrain the combination of the eyeball mesh and the neural SDF field.
Such a hybrid representation not only helps us to bypass the challenge of reconstructing the highly reflective eyeballs but also makes our method fully compatible with the graphics pipeline.

\section{Method}
In this Section, we first introduce our capture setup (Section~\ref{sec:method:data}). 
We then propose a novel hybrid representation for high-quality and complete face modeling (Section~\ref{sec:method:hybrid-rep}).
To train it from the captured data, we propose a novel mesh-aware volume rendering technique (Section~\ref{sec:method:mesh-aware-render}) and a set of carefully-designed training strategies (Section~\ref{sec:method:train}).

\subsection{Data Capture}
\label{sec:method:data}
As illustrated in Figure~\ref{Fig:teaser}, we capture a single video sequence around the subject in a dim room using the smartphone camera with its flashlight opened up.
The capture takes around 25 seconds for a subject.
We use an iPhone X to capture all the sequences in this paper.
We resize the images to $960\times 720$ resolution before processing.
We calibrate the camera parameters for each frame using an off-the-shelf software\footnote{\href{https://www.agisoft.com/}{https://www.agisoft.com/}}.
We assume the only high-frequency light source is the smartphone flashlight and further assume it shares the same position as the camera; if not otherwise specified, the data is captured in a room at night\footnote{It is not a darkroom as the white wall and furniture would reflect light.}.
Such a setup has several advantages: \emph{i}) it is easy to fulfill for daily users at home, \emph{ii}) the high-frequency flashlight provides rich cues to recover reflectance, and \emph{iii}) there are no apparent shadows in the captured frames, avoiding extra efforts during optimization. 

\subsection{Hybrid Representation}
\label{sec:method:hybrid-rep}
Our goal is to capture the complete face with skin, hair, mouth interior, and eyes from the recorded frames.
In addition, we require the resulting assets to be compatible with common CG software.
In this context, neural fields become the first choice for their superior representation power and flexibility.
Similar to existing works~\cite{zhang2022iron,cheng2023wildlight}, we can represent geometry as the neural SDF field and adopt a neural field to model the BRDF parameters as the reflectance; at test time, we can export it to 3D assets including a mesh and a set of UV maps.
However, such a holistic representation fails to capture plausible geometry and reflectance for the highly reflective eyeballs, leading to unpleasing results around the eyes (see Figure~\ref{Fig:exp-hybrid} and~\ref{Fig:exp-hybrid-relight}).
To overcome the challenging problem of eyeball reconstruction, we propose a hybrid representation to explicitly exploit eyeball priors.
Specifically, we split the whole face $F$ into the eyeballs region $E$ and all the other regions $S$, \emph{i.e.} $F\!\!=\!\!E\cup S$.

For the $S$ region, including skin, mouth interior, and hair, we adopt a neural SDF field $f_{sdf}: \mathbf{x} \rightarrow sdf_{S}$ to represent geometry and a neural field $f_{brdf}: \mathbf{x} \rightarrow \{c,s_S, \rho_S\}$ to model the BRDF parameters as the reflectance.
Here, $\mathbf{x}\in\mathbb{R}^3$ is the position of the sampled point; $c\in\mathbb{R}^3$, $s_S\in\mathbb{R}$, and $\rho_S\in\mathbb{R}$ are the diffuse albedo, specular albedo, and roughness of the Disney BRDF model~\cite{burley2012physically} respectively.

For eyeballs, \emph{i.e.} the $E$ region, we make an assumption that they are two sphere meshes with the same radius. 
We further assign a specular lobe with predefined specular albedo $s_{E}$ and roughness $\rho_{E}$ to the eyeballs, while leaving the spatially varying diffuse albedo of the eyeballs to be solved from the recorded data.
Thus, the only person-specific characteristics of the eyeballs are the two positions $\mathbf{p}_{l},\mathbf{p}_{r}\in\mathbb{R}^3$, a shared scalar radius $r\in\mathbb{R}$, and the diffuse albedo.
We reuse $f_{brdf}$ to represent the spatially varying diffuse albedo of the two eyeballs.
Although simple, we demonstrate decent rendering results with plausible specularities appearing on the eyes.

\subsection{Mesh-Aware Volume Rendering}
\label{sec:method:mesh-aware-render}

\begin{figure}[t]
    \centering
    \includegraphics[width=0.45\textwidth]{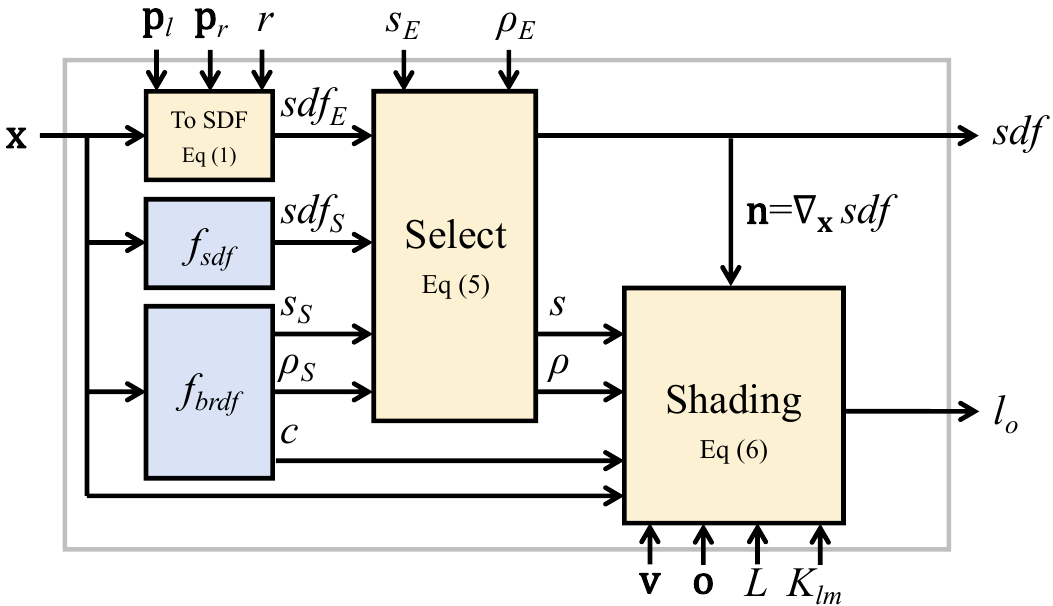}
    \vspace{-5pt}
    \caption{Illustration of the proposed mesh-aware volume rendering technique tailored to our novel hybrid face representation. }
    \label{Fig:vol-render}
\end{figure}

To train our hybrid representation, we need to render it into an image to compute the photometric loss against the recorded frame. 
To this end, we propose a mesh-aware volume rendering technique tailored to our hybrid representation as illustrated in Figure~\ref{Fig:vol-render}.
Specifically, given a 3D position $\mathbf{x}=\mathbf{o}+t\cdot\mathbf{d}$ sampled on the camera ray where $\mathbf{o}\in\mathbb{R}^3$ is the camera position, $\mathbf{d}\in\mathbb{R}^3$ is the opposite view direction, and $t\in\mathbb{R}$ is the viewing distance, we introduce how to compute its SDF value, normal, and observed color. 

We first convert the eyeball meshes to the SDF field:
\begin{equation}
\label{eq:sdf-sphere}
    sdf_{E} = \min (||\mathbf{x}-\mathbf{p}_{l}||_2 - r, ||\mathbf{x}-\mathbf{p}_{r}||_2 - r )
\end{equation}
Considering that the complete facial region $F$ is the union of the eyeballs region $E$ and the other region $S$, we compute its SDF value $sdf$, specular albedo $s$, and roughness $\rho$ as: 
\begin{align}
    s &= {\rm select}(s_E, s_S; sdf_E, sdf_S) \\
    \rho &= {\rm select}(\rho_E, \rho_S; sdf_E, sdf_S) \\
    sdf &= {\rm select}(sdf_E, sdf_S; sdf_E, sdf_S) \label{eq:sel_sdf}
\end{align}
The intuition is that we select its attributes modeled by either the $E$ region or the $S$ region according to its geometry. 
We define the differentiable select operator as:
\begin{equation}
    {\rm select}(*_E,*_S;sdf_E,sdf_S) = 
    \begin{cases}
        *_E & sdf_E\leq sdf_S \\
        *_S & sdf_E > sdf_S
    \end{cases}
\end{equation}
Note that Eq~\eqref{eq:sel_sdf} is equivalent to setting the SDF value of the union region $F$ as the minimum SDF value of its two components, \emph{i.e.} $E$ and $S$. 
The diffuse albedo of region $F$ is directly set to $c$ considering that we use $f_{brdf}$ to represent both regions.
We compute the normal $\mathbf{n}\in\mathbb{R}^3$ as the gradient of the SDF value \emph{w.r.t} the position, \emph{i.e.} $\mathbf{n}=\nabla_{\mathbf{x}} sdf$.

\textbf{Combined Lighting Model.} 
We represent lighting in our capture setup as a combination of the high-frequency smartphone flashlight and the low-frequency dim ambient light.
We parametrize the flashlight as a point light with predefined 3-channel intensity $L$.
For the ambient light, we only consider its contribution to the diffuse term; we parameterize the diffuse shading under the ambient light as the 2-order Spherical Harmonics (SH)~\cite{ramamoorthi2001efficient} in the $\rm SoftPlus$ output space to ensure its non-negativity.
Then, given the material parameters and the normal, we can compute its shading as:
\begin{align}
    l_o &= l_{flash} + l_{amb} {\rm, where} \\
    l_{flash} &= \frac{L}{||\mathbf{x}-\mathbf{o}||_2^2}\cdot f_{pbr}(\mathbf{l},\mathbf{v};c,s,\rho)\cdot\max(\mathbf{n}\cdot\mathbf{v},0) \\
    l_{amb} &= c\cdot {\rm SoftPlus}(\sum_{l=0}^2\sum_{m=-l}^{l}\cdot K_{lm}\cdot Y_{lm}(\textbf{n}))
\end{align}
Here, $f_{pbr}$ is the Disney BRDF, $\mathbf{v}=-\mathbf{d}$ is the view direction, $\mathbf{l}$ is the light direction and we have $\mathbf{l}=\mathbf{v}$ in the co-located setup, $K_{lm}$ are the SH coefficients for the diffuse shading under the ambient light, and $Y_{lm}(\cdot)$ are the SH bases.
Compared to WildLight~\cite{cheng2023wildlight} which uses NeRF~\cite{mildenhall2020nerf} to represent the ambient shading, our representation is more compact.
Thus, in our scenario, it is feasible to estimate it from a single flashlight video.
In addition, it makes our capture process faster, enabling capturing more challenging facial expressions.

Given the SDF value $sdf$, normal $\mathbf{n}$, and the observed color $l_o$ of a sample point $\mathbf{x}$, we can volume render the camera ray following VolSDF~\cite{wang2021neus}.

\subsection{Training}
\label{sec:method:train}
We learn the two neural fields $f_{sdf}$ and $f_{brdf}$ and the ambient shading parameters $K_{lm}$ from the captured frames.
Similar to EyeNeRF~\cite{li2022eyenerf}, the eyeball position $\mathbf{p}_{l},\mathbf{p}_{r}$, and radius $r$ are set manually, which can be easily done in CG software like Blender.
We leave incorporating automatic method~\cite{wen2020accurate} into our system as the future work.
We emphasize that our goal is to exploit priors to overcome the challenge problem of reconstructing the highly reflective eyeballs, rather than reconstructing high-quality eyeballs with accurate positions and characteristics~\cite{berard2014high}.

\begin{figure}[t]
    \centering
    \includegraphics[width=0.475\textwidth]{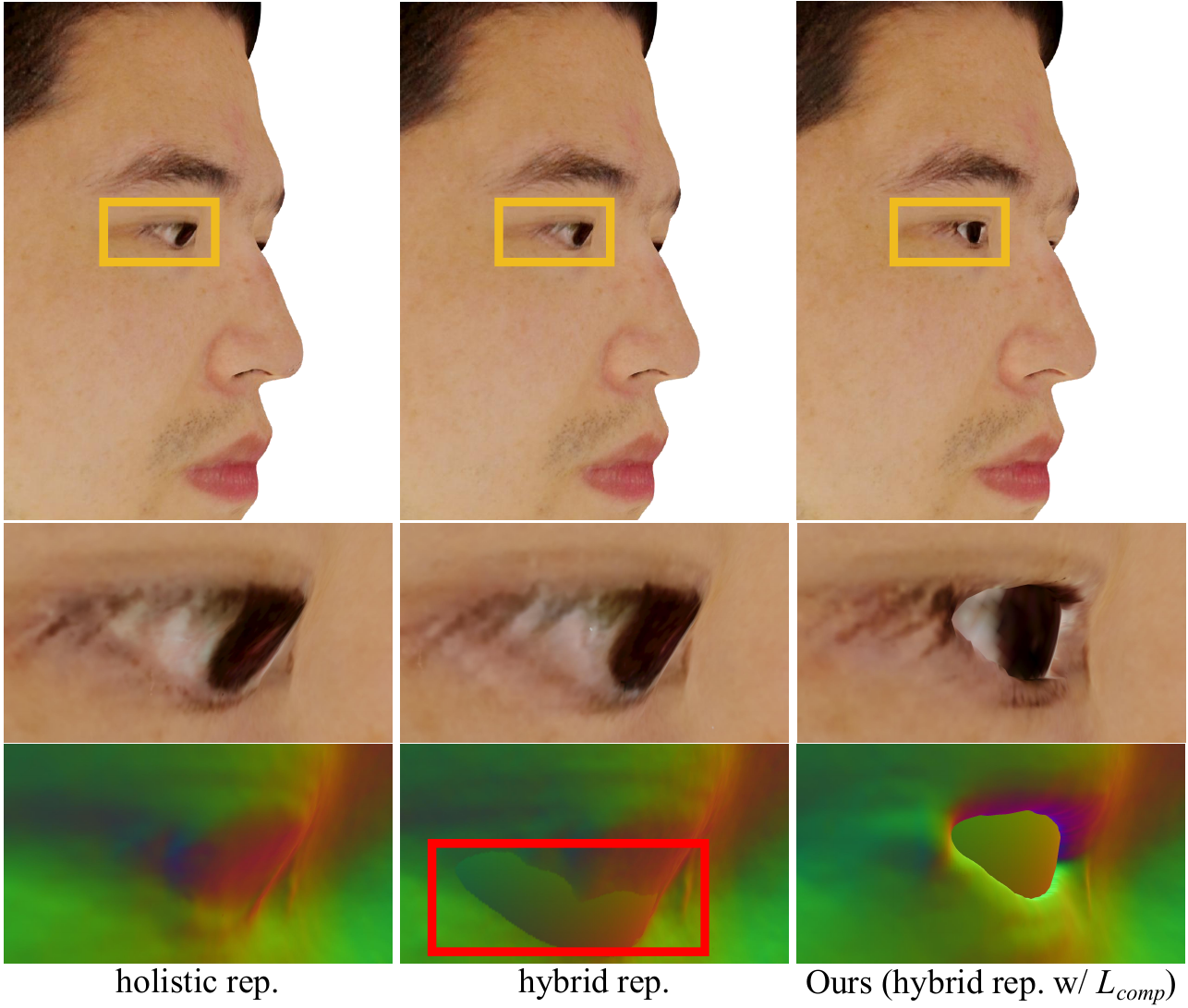}
    \vspace{-15pt}
    \caption{Qualitative evaluation of the hybrid representation and $\mathcal{L}_{comp}$ on geometry reconstruction around eyes. Texture and normal close-ups are shown in the second and third rows respectively. }
    \label{Fig:exp-hybrid}
\end{figure}

To train the neural fields from images, we adopt photometric loss, mask loss, and Eikonal loss~\cite{icml2020_2086} similar to previous works~\cite{cheng2023wildlight}.
We also adopt a loss to enforce the normal of the nearby sampled points to be the same~\cite{zhang2021nerfactor,rosu2023permutosdf}.
See more details in our \emph{supplementary material}.
However, all these routine loss functions provide no explicit constraints to regularize the eyeball meshes and the neural SDF field to only represent their own region.
Thus, it leads to unnatural results as shown in Figure~\ref{Fig:exp-hybrid}.
In addition, our method focuses on capturing the face rather than common objects so that we can exploit face-specific priors for regularization. 
To this end, we propose two novel and well-designed losses tailored to our task and hybrid representation.

\textbf{Composition Loss.}
To constrain the training of our hybrid representation, inspired by ObjectSDF++~\cite{wu2023objectsdf++}, we render the occlusion-aware object opacity mask $\hat{O}^{E}$ and $\hat{O}^{S}$ for the $E$ and $S$ region and compare them to the corresponding ground truth ${O}^{E}$ and ${O}^{S}$ obtained from an off-the-shelf face parsing network~\cite{lin2021roi} over the $n$ sampled rays:
\begin{equation}
    \mathcal{L}_{comp} = \sum_{i=1}^n||\hat{O}^{E}_i - O^{E}_i||_1 + \sum_{i=1}^n||\hat{O}^{S}_i - O^{S}_i||_1
\end{equation}

\textbf{Reflectance Regularization.}
We exploit the morphable face albedo model -- AlbedoMM~\cite{smith2020morphable} -- as the reflectance prior.
Specifically, we devise a multi-view AlbedoMM fitting algorithm to reconstruct the specular albedo for each frame.
Then, we enlarge the solved specular albedo to the whole image to obtain $I^s$ as pseudo ground truth to supervise the volume-rendered one $\hat{I}^s$ on the sampled rays:
\begin{equation}
    \mathcal{L}_{ref} = \sum_{i=1}^n||k\cdot \hat{I}^s_i - I^s_i||_1
\end{equation}
Here, $k\in\mathbb{R}$ is a learnable scalar to compensate for the scale ambiguity stemming from our predefined light intensity $L$.
For pixels from the eyeballs region $E$, we do not compute $\mathcal{L}_{ref}$ since we already have predefined prior $s_{eye}$. 
For pixels from the hair region indicated by the parsing mask~\cite{lin2021roi}, we constrain its specular albedo to be $0$ to obtain a diffuse appearance as we empirically find fitting a specular lobe produces artifacts when rendered in novel environments.

\section{Experiments}
We implement our method on top of the multi-resolution hash grid~\cite{muller2022instant} and VolSDF~\cite{yariv2021volume} using NerfAcc~\cite{li2022nerfacc}.
Our method can be trained within 70 minutes using a single Nvidia RTX 3090 graphics card.
After training, we automatically export our hybrid representation to a triangle mesh with corresponding UV maps for normal, diffuse albedo, specular albedo, and roughness as shown in Figure~\ref{Fig:teaser}.
We adopt Blender to re-render these assets in novel environments.
We urge the readers to check our \emph{supplementary video} and \emph{supplementary material} for more implementation details, experimental results, and illustrations.

\subsection{Evaluations}

\begin{figure}[t]
    \centering
    \includegraphics[width=0.475\textwidth]{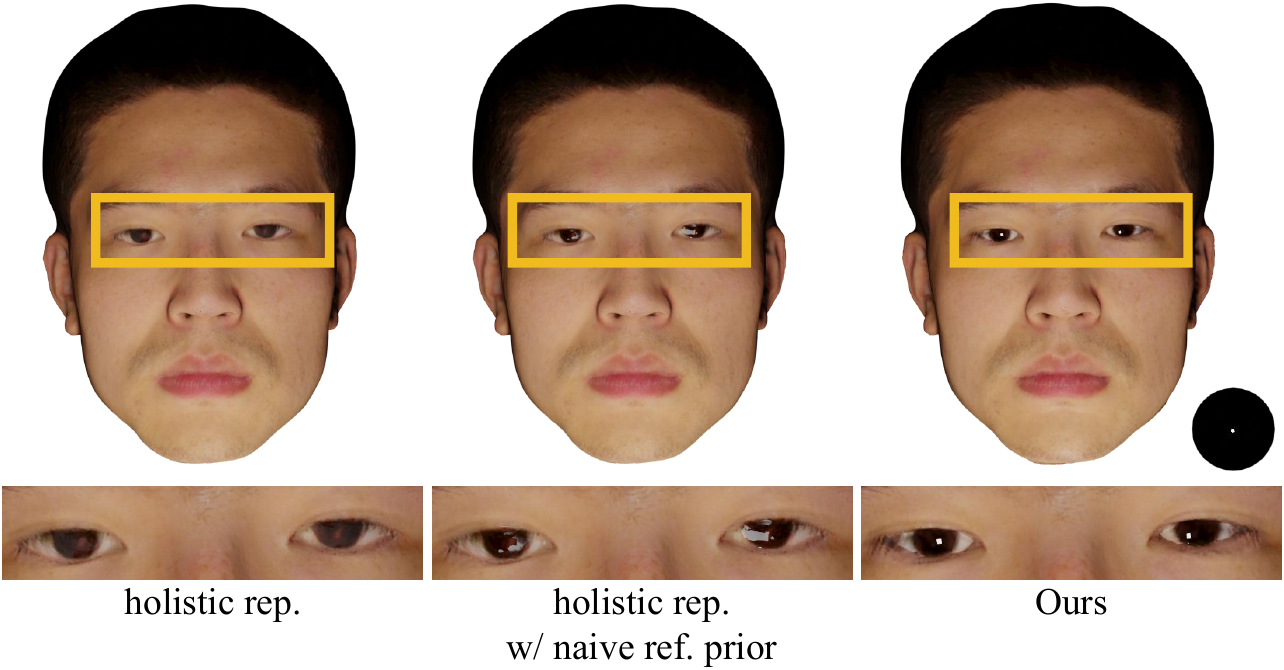}
    \vspace{-15pt}
    \caption{Qualitative evaluation of our hybrid representation and the baseline variants on relighting under a frontal point light. }
    \label{Fig:exp-hybrid-relight}
\end{figure}

\textbf{Hybrid Face Representation and $\mathcal{L}_{comp}$.}
Recall that our motivation for proposing the hybrid face representation is to alleviate the challenging problem of reconstructing the highly reflective eyeballs' geometry and appearance from images.
To evaluate its effectiveness, we compare a baseline variant \emph{holistic rep.}, where the whole facial geometry and reflectance are represented by $f_{sdf}$ and $f_{brdf}$.
In addition, we compare to a baseline variant \emph{hybrid rep.} where we remove $\mathcal{L}_{comp}$ from our full method to evaluate its efficacy.

We show the geometry reconstruction results in Figure~\ref{Fig:exp-hybrid}.
Without eyeballs prior, \emph{holistic rep.} fails to reconstruct reasonable eyeball geometry.
Without the composition loss $\mathcal{L}_{comp}$, we cannot ensure the eyeball meshes and the neural SDF represent their own region as we expected. 
Our method, \emph{i.e. hybrid rep. w/ $\mathcal{L}_{comp}$} obtains the best results.
It seamlessly integrates the eyeballs' geometry and reflectance prior to the hybrid face representation and constrains its learning via the composition loss $\mathcal{L}_{comp}$.

We show the relighting results in Figure~\ref{Fig:exp-hybrid-relight}.
The baseline variant \emph{holistic rep.} models eyeballs' reflectance the same way as the other facial regions.
It fails to reconstruct a plausible specular lobe for eyeballs, leading to unnatural diffuse-looking relighting results around the eyes.
We further enhance it by manually setting the specular albedo and roughness of the eyeballs region on the exported UV maps to $s_{eye}$ and $\rho_{eye}$ respectively; we dub this one \emph{holistic rep. w/ naive ref. prior}.
Although reflectance prior is utilized, it still cannot generate plausible specularities due to the erroneous geometry reconstruction.
Our method produces plausible specularities in eye renderings since we exploit both geometry and reflectance eyeball priors.

\textbf{Combined Light Representation.}
To evaluate the effectiveness of our combined light representation, we compare a baseline variant that only uses a point light to represent the dominant flashlight while ignoring the ambient.
We report the performance gain (in terms of PSNR) on face reconstruction of our method over this baseline on $3$ capture environments with increasing ambient intensity: $0.08$dB for \emph{night w/ curtain}, $0.05$dB for \emph{noon w/ curtain}, and $0.61$dB for \emph{noon w/o curtain}.
See the photo of these environments and more evaluations in \emph{supplementary material}.

\begin{figure}[t]
    \centering
    \includegraphics[width=0.475\textwidth]{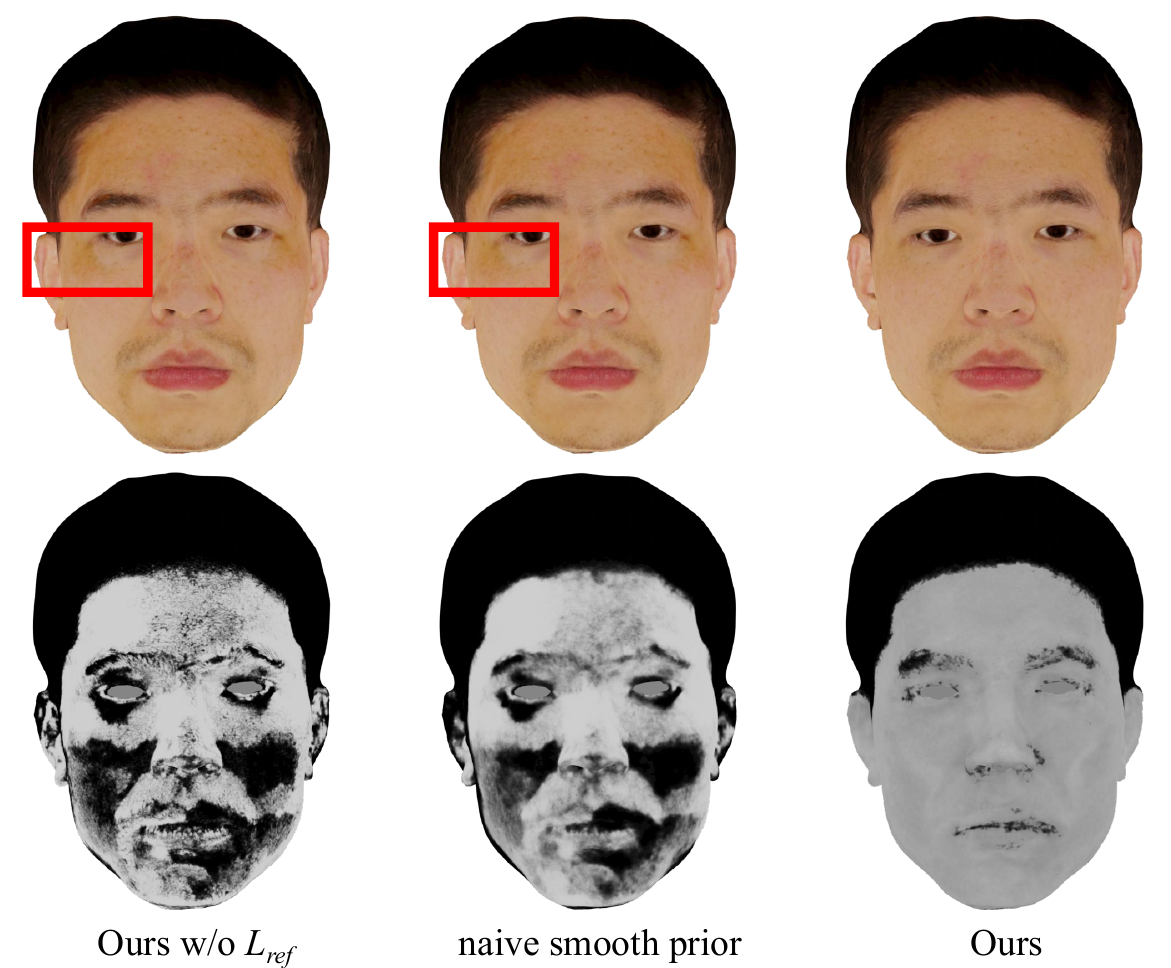}
    \vspace{-15pt}
    \caption{Qualitative evaluation of our reflectance regularization loss $\mathcal{L}_{ref}$ and the baseline variants on diffuse (the first row) and specular (the second row) albedo estimation.
    See our \emph{supplementary video} for more illustrations. }
    \label{Fig:exp-abdmm}
\end{figure}

\textbf{Reflectance Regularization.}
To regularize the estimated reflectance, we exploit AlbedoMM~\cite{smith2020morphable} as priors.
We qualitatively evaluate this term as it is a regularizer.
As shown in Figure~\ref{Fig:exp-abdmm}, without our reflectance prior loss $\mathcal{L}_{ref}$, the estimated diffuse and specular albedo have a degraded quality compared to the full method.
We also compare a widely-used smooth prior that constrains the nearby points' estimated specular albedo to be the same~\cite{zhang2021nerfactor,Munkberg_2022_CVPR,hasselgren2022shape} in Figure~\ref{Fig:exp-abdmm}. 
Again, our method obtains superior quality on diffuse and specular albedo estimation over this naive prior.

\subsection{Comparisons}
We compare state-of-the-art inverse rendering methods in various problem setups, from low-cost systems to studios.
The first class of works takes multi-view images captured in an unconstrained environment as input, which is a bit easier to use than our method; we compare the latest one, \emph{i.e.} NeRO~\cite{liu2023nero}, as the representative work.
The second class takes the same input as our method; we involve a model-based method NextFace++ for comparison.
We also test WildLight~\cite{cheng2023wildlight} under this setup.
The last class is studio-based methods; we involve a Light Stage-based solution~\cite{ghosh2011multiview}.
Due to space limitation, we put the comparison to NeRO and WildLight in our \emph{supplementary material}.
We do not compare to PolFace~\cite{azinovic2023high} as it is closed-source.

\begin{figure}[t]
    \centering
    \includegraphics[width=0.475\textwidth]{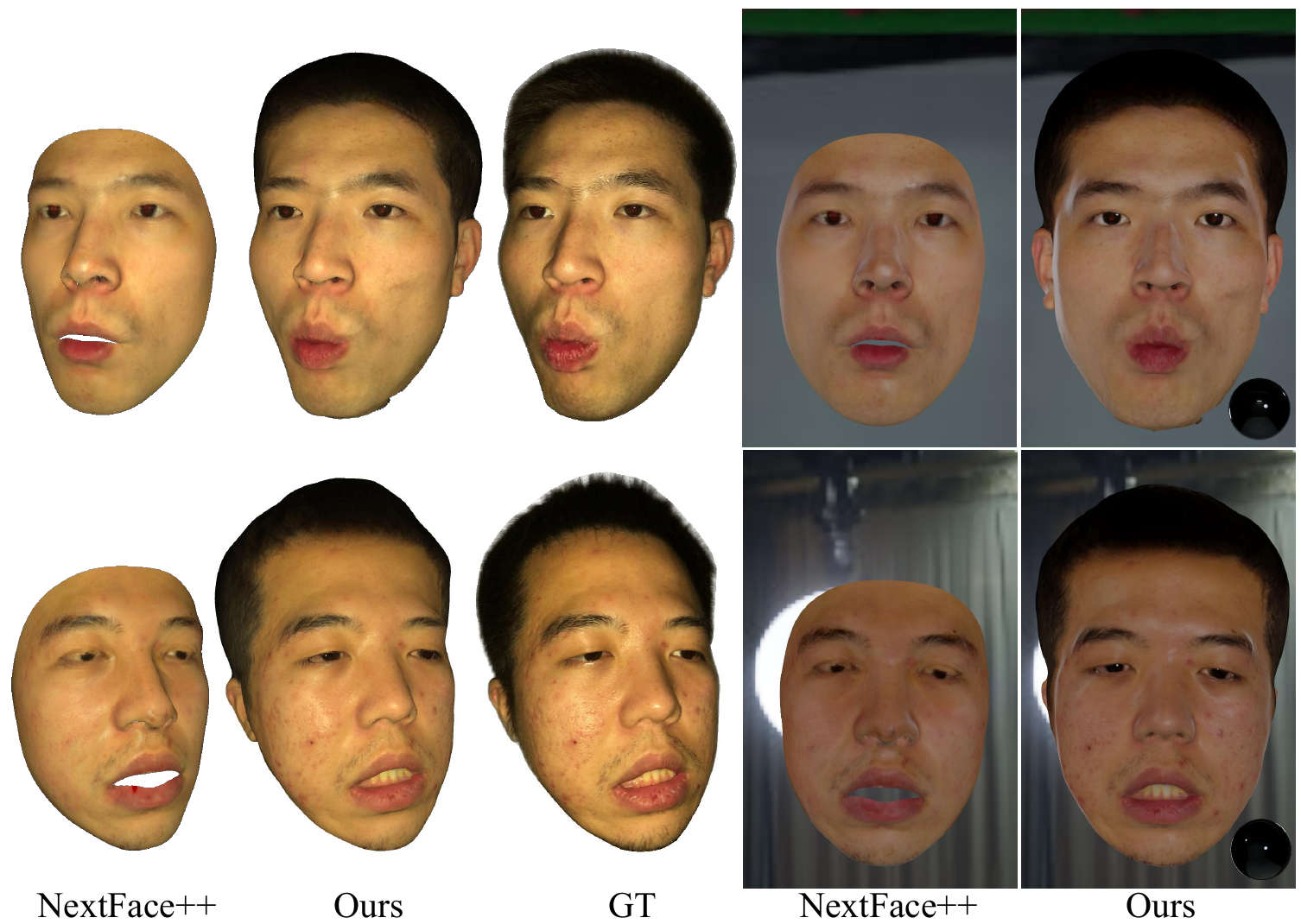}
    \vspace{-15pt}
    \caption{Qualitative comparison of our method and NextFace++ on face reconstruction and relighting. }
    \label{Fig:exp-nextface}
\end{figure}

\begin{table}[t]
\centering
\begin{tabular}{@{}lllll@{}}
\toprule
 & PSNR~$\uparrow$ & SSIM~\cite{wang2004image}~$\uparrow$ & 
 LPIPS~\cite{zhang2018perceptual}~$\downarrow$  \\ \midrule
 NextFace++ & 17.62 & 0.7339 & 0.2727 \\ \midrule
 Ours & \textbf{26.12} & \textbf{0.8808} & \textbf{0.1642}  \\ \bottomrule
\end{tabular}
\caption{Quantitative comparison of our method and NextFace++ on face reconstruction. The metric is averaged on $5$ subjects.}
\label{tab:cmp}
\end{table}

\textbf{Comparison to NextFace++.}
NextFace~\cite{dib2021practical} takes single or multiple in-the-wild face images as input.
It first fits 3DMM to the images by estimating the lighting, camera parameters, head pose, BFM geometry parameter~\cite{gerig2018morphable}, and AlbedoMM reflectance parameter~\cite{smith2020morphable}.
Then, they refine the statistical reflectance maps on a per-texel basis.
For a fair comparison, we enhance NextFace in the following aspects: (1) we provide our camera parameters to NextFace and introduce a learnable 1D scalar to compensate for the scale difference between the BFM canonical space and our camera frame, and (2) we implement our combined lighting model in NextFace.
We dub it NextFace++.

We compare the face reconstruction and relighting results of our method to NextFace++.
As shown in Figure~\ref{Fig:exp-nextface}, NextFace++ can only represent facial skin as it relies on the BFM geometry while our method can represent the complete face thanks to the proposed hybrid representation.
On face reconstruction, NextFace++ is confined to the space of the BFM model, which cannot represent person-specific characteristics around the eyes, nose, and mouth, while our method can better fit the captured images as our hybrid geometry representation is more powerful and flexible.
On face relighting, our method achieves more realistic results around eyes as we exploit eyeballs' reflectance prior while NextFace++ models them the same way as skin.
In Table~\ref{tab:cmp}, we show quantitative metrics on the hold-out validation images sampled from the captured co-located sequence; we obtain superior results over NextFace++.

\begin{figure}[t]
    \centering
    \includegraphics[width=0.475\textwidth]{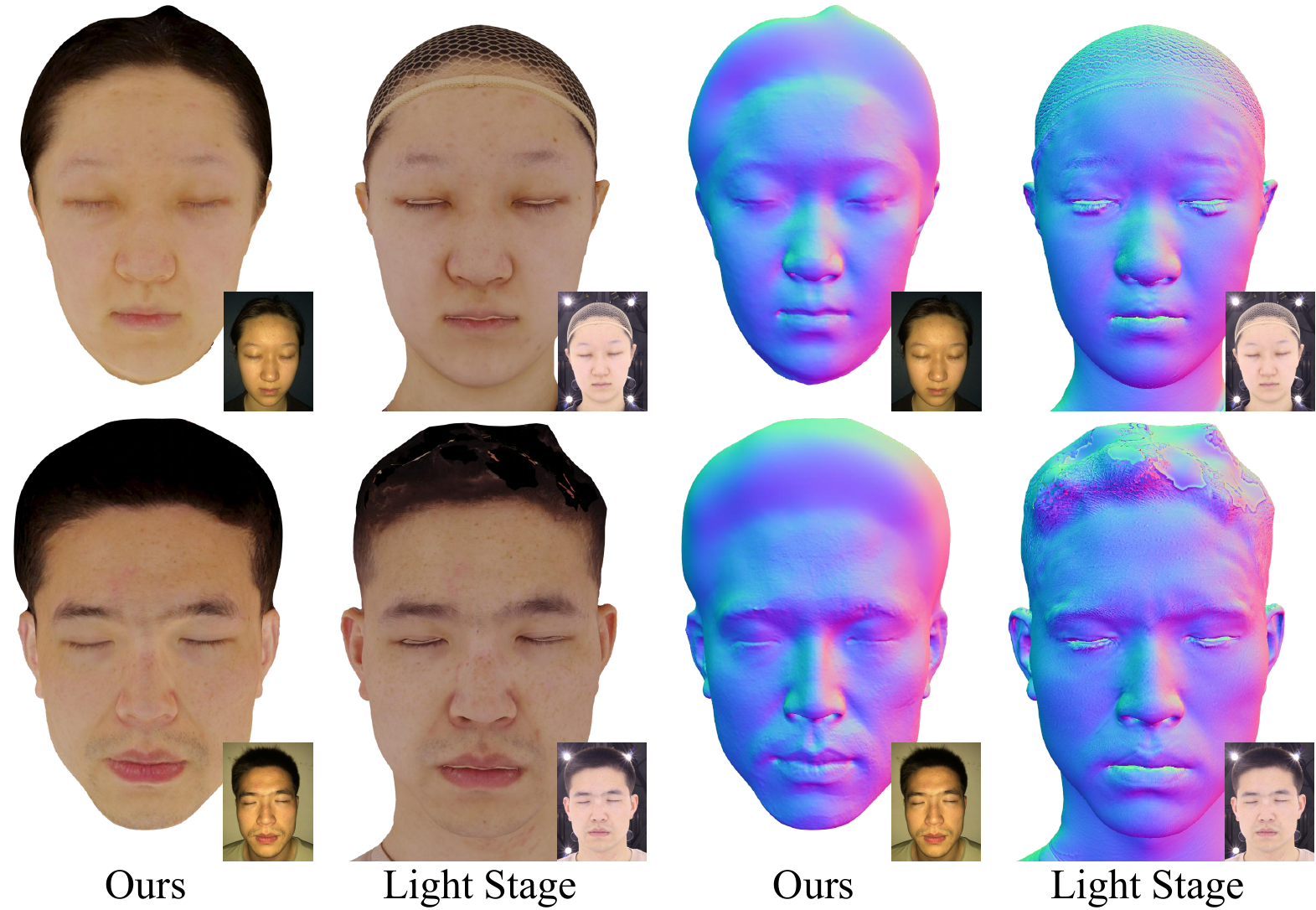}
    \vspace{-15pt}
    \caption{Qualitative comparison of our method and the Light Stage-based solution~\cite{ghosh2011multiview} on diffuse albedo and normal reconstruction. 
    We show one frontal reference frame sampled from the recorded data at the right bottom corner of each image.
    }
    \label{Fig:exp-lightstage}
\end{figure}

\textbf{Comparison to Light Stage.}
In high-budget production, Light Stage~\cite{debevec2012light} has demonstrated tremendous success in 3D scanning real humans~\cite{alexander2009digital,alexander2013digital}.
To evaluate the performance gap between our low-cost method and the state-of-the-art in the studio, we compare it to the Light Stage-based solution~\cite{ghosh2011multiview} implemented by a company\footnote{\href{http://soulshell.cn/}{http://soulshell.cn/}}
In their Light Stage, polarization filters are equipped on lights and cameras to capture the diffuse albedo while sphere gradient illuminations are activated to capture the normal; their system cannot capture specular albedo currently.
We invite volunteers to their studio for on-site capture.

In Figure~\ref{Fig:exp-lightstage}, we compare the diffuse albedo and normal reconstructed from our method and the Light Stage.
Although impressive results are achieved, our method still legs behind the Light Stage results in several aspects: (1) Light Stage can better disentangle the diffuse and specular components due to its usage of the polarization filters, leading to a cleaner diffuse albedo map, and (2) Light Stage can reconstruct higher resolution maps with pore level details since it uses high-definition DSLR camera to capture data, while our method is limited by the quality of the smartphone video camera and the inevitable subtle movement of the subject during our longer capture process (around 25 seconds).

\begin{figure}[t]
    \centering
    \includegraphics[width=0.475\textwidth]{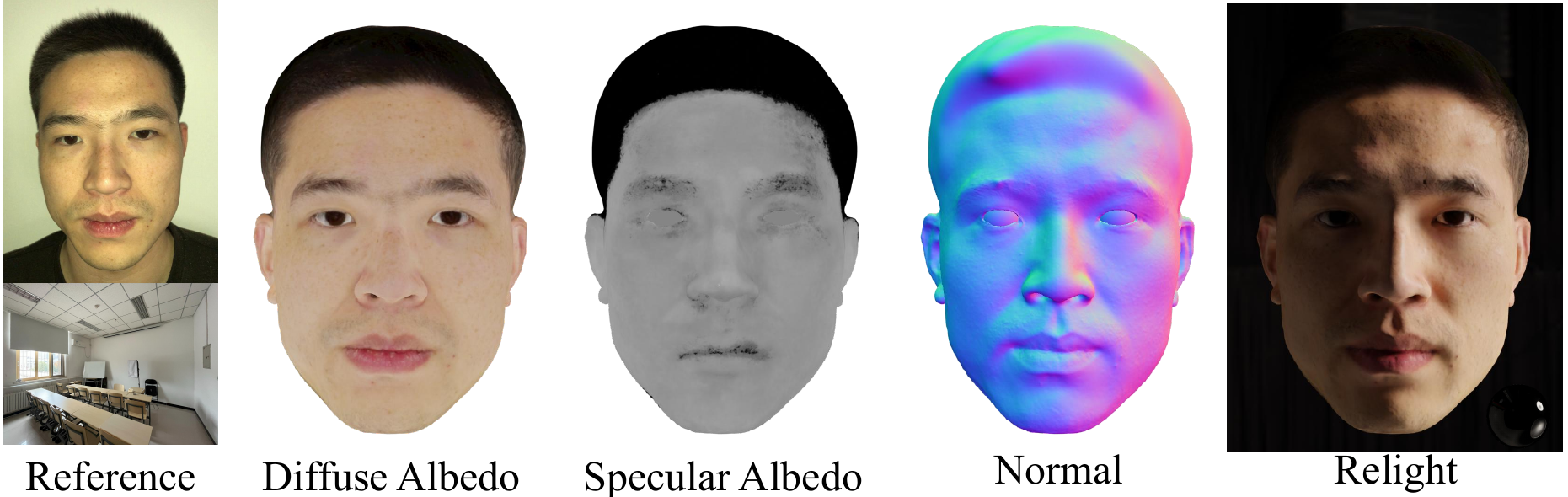}
    \vspace{-15pt}
    \caption{Our method can reconstruct high-quality facial geometry and reflectance even if apparent ambient exists. We show a frontal view sampled from the recorded video and the photo of the scene where we capture data (\emph{noon w/o curtain}) in the leftmost column.}
    \label{Fig:exp-robust}
\end{figure}

\begin{figure}[t]
    \centering
    \includegraphics[width=0.475\textwidth]{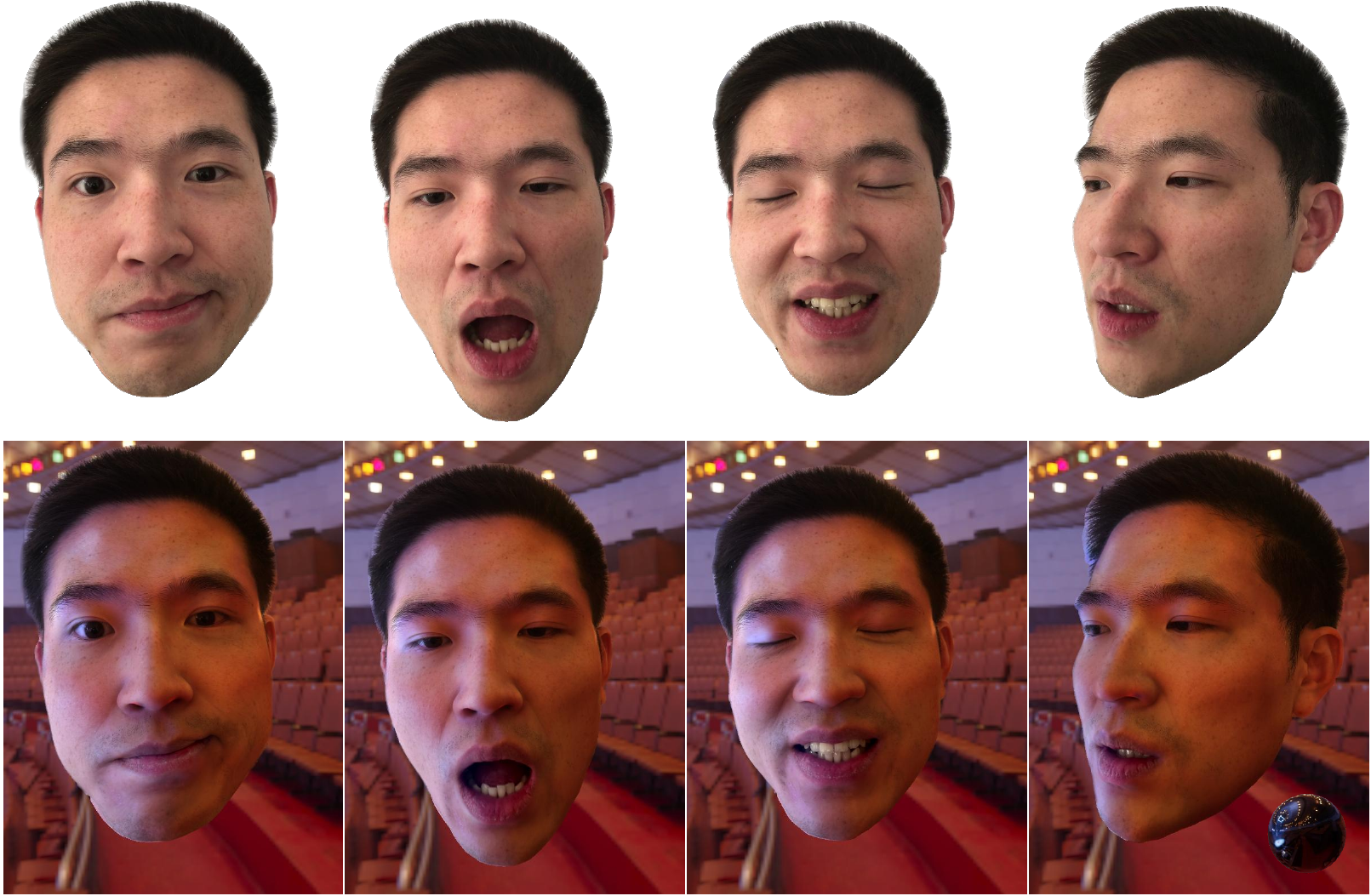}
    \vspace{-15pt}
    \caption{Qualitative face performance relighting results obtained by combining our method with the Reflectance Transfer~\cite{peers2007post}. We show the origin performance sequence and the relit one in the first and second row respectively. }
    \label{Fig:exp-ref-transfer}
\end{figure}

\begin{figure*}[t]
    \centering
    \includegraphics[width=\textwidth]{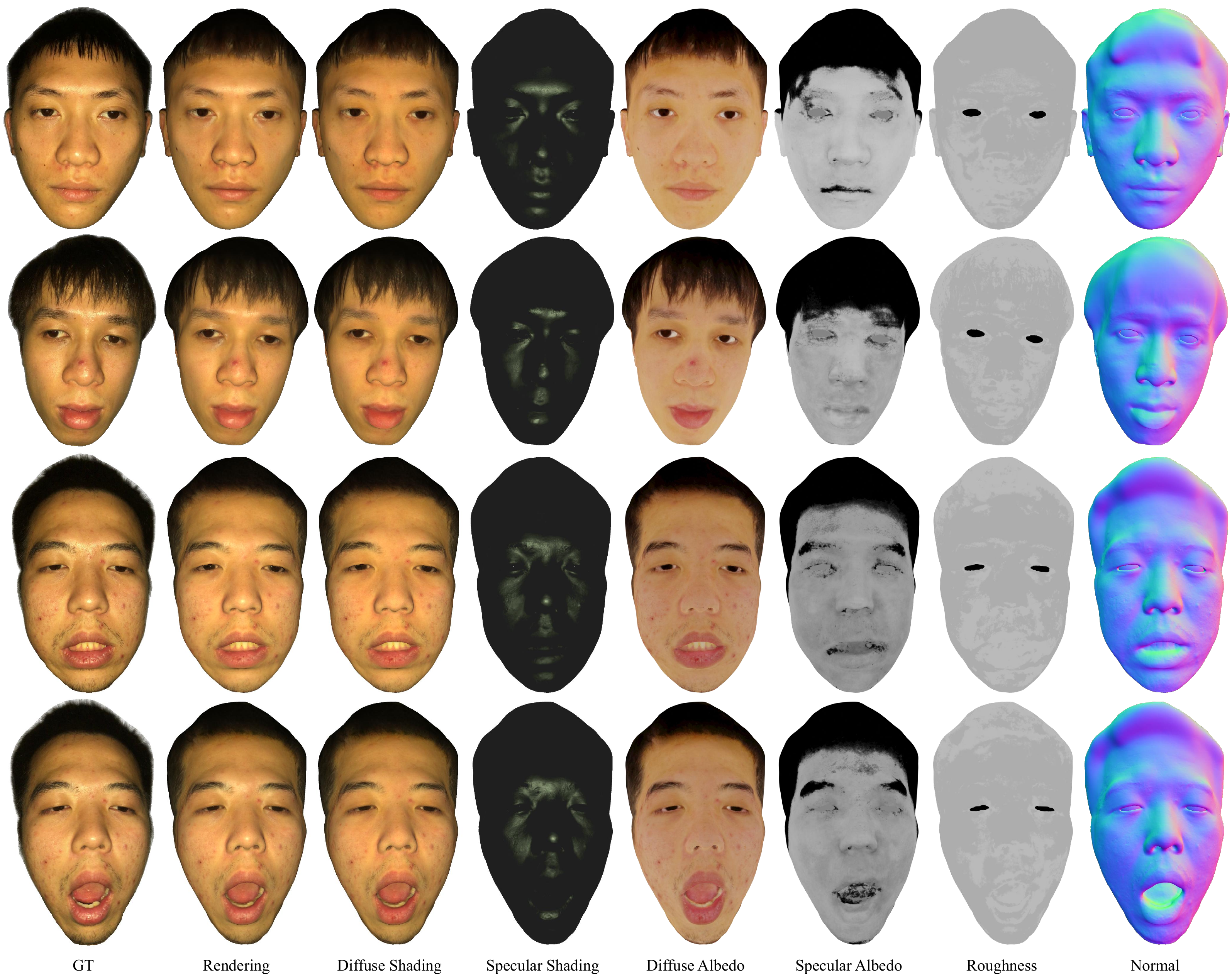}
    \vspace{-15pt}
    \caption{Facial geometry and appearance capture results of our method on different identities and facial expressions. }
    \label{Fig:exp-rerender}
\end{figure*}

\subsection{Results and Application}
We present the face capture results of our method on different identities and facial expressions in Figure~\ref{Fig:exp-rerender}.
Our method can disentangle the diffuse and specular components from the images in a plausible way, leading to an authentic high-quality relightable scan.
In addition, our method can capture various challenging facial expressions thanks to the strong representation power and flexibility of our hybrid face representation.
In Figure~\ref{Fig:exp-robust}, we demonstrate the robustness of our method by training it from the data captured at noon in a room with the curtain opened, \emph{i.e. noon w/o curtain}.
In this challenging scenario with apparent ambient, our method still obtains high-quality results.

By replacing the Light Stage scan in the Reflectance Transfer technique~\cite{peers2007post} as our method's result, we build a simple but strong baseline for the challenging task of relightable face performance capture in the low-cost setup.
We record a performance sequence under an unknown but low-frequency lighting and make it relightable as shown in Figure~\ref{Fig:exp-ref-transfer}.
See the \emph{supplementary material} for more details.

\subsection{Limitations and Discussions}
Although our method demonstrates high-quality results in an easy-to-use manner, it still has several limitations.
Similar to EyeNeRF~\cite{li2022eyenerf}, the position and radius of the eyeball meshes are manually set in our method, which incurs some manual effort to the whole pipeline.
Pre-capturing a multi-gaze video to automatically estimate the eyeball position and size~\cite{wen2020accurate} is an interesting direction.
The eye rendering quality is limited by the reflectance and geometry priors we use.
Incorporating a more complex model~\cite{berard2014high} is a promising future work.
To fulfill this, we can modify Equation~\eqref{eq:sdf-sphere} to support general eyeball meshes; note that our mesh-aware volume rendering can be extended to arbitrary mesh geometry naturally. 
We observe eyelashes baked in the eyeball's texture as we do not explicitly model them.
However, this artifact is not apparent on the whole face's scale.
Our method takes around 25 seconds to capture a subject with a fixed facial expression.
During this period, subtle movement of the subject is inevitable, which would blur the reconstructed texture or bake the eyelids into the eyeball's texture.
Speeding up the capture process is an interesting direction.
We currently assume hair is a diffuse surface.
We leave integrating strand-based hair prior to our hybrid representation as our future work.
See our \emph{supplementary material} for more detailed discussions.

\section{Conclusion}
We propose a low-cost and easy-to-use technique for high-quality facial geometry and appearance capture, which takes a single co-located smartphone flashlight sequence captured in a dim room as input.
Our method can model the complete face with skin, hair, mouth interior, and eyes.
We propose a novel hybrid face representation by combining meshes and neural SDF field and techniques to train it from images.
We apply a combined lighting model to compactly model the illumination and propose to exploit AlbedoMM~\cite{smith2020morphable} as priors to constrain the estimated reflectance.
Our method reconstructs high-quality 3D relightable scans compatible with common CG software.

\clearpage
\maketitlesupplementary
\appendix

\section{More Implementation Details}
We provide some important implementation details in the following.
The full training code and the data processing code will be released in the future.

\subsection{Camera Calibration}
\label{sec:supp:cam_calib}
We adopt MetaShape\footnote{\href{https://www.agisoft.com/}{https://www.agisoft.com/}} to calibrate a shared camera intrinsic matrix for all the recorded frames and a camera extrinsic matrix for each frame.
We translate and scale the mesh reconstructed by MetaShape to fit it into the $[-1,1]^3$ cube; the camera extrinsic matrices are transformed the same way as the mesh.
In this way, we ensure that a $[-1,1]^3$ bounding box is enough for the neural field to represent the whole face.
During the data capture process, we set the ISO to 300, the white balance to 4900K, and the FPS to 30. 

\subsection{Smartphone Flashlight Calibration}
Recall that we parameterize the smartphone flashlight as a point light source with 3-channel intensity $L$.
We further represent $L$ as the multiplication of a 1-channel scale $s_L\in\mathbb{R}$ and a 3-channel RGB color $c_L\in\mathbb{R}^3$, \emph{i.e.} $L=s_L\cdot c_L$.
We calibrate $c_L$ by capturing a smartphone flashlight image for a pure-white A4 page. 
We then adopt the mean color of a select patch on this image as $c_L$.
We empirically set $s_L=8$ and find it works well for all the subjects we captured following the camera calibration procedure in Section~\ref{sec:supp:cam_calib}.

\subsection{Disney BRDF Implementation}
We adopt a modified version of the Disney BRDF $f_{pbr}$~\cite{burley2012physically}, containing a diffuse term and a specular term.
Both terms are implemented identically to WildLight~\cite{cheng2023wildlight}; see their paper and open-source code for more details.

\subsection{Network Architecture}
We implement our neural field on top of the multi-resolution hash grid~\cite{muller2022instant}.
The neural SDF field and the neural reflectance field are implemented as independent hash grids.
The neural SDF field is initialized to a sphere to stabilize training~\cite{icml2020_2086}.
In addition, we add a small MLP head on the neural SDF field to predict the view-dependent color.

During training, the photometric loss is computed for both the predicted one (\emph{i.e.} the view-dependent color predicted by the small MLP head) and the physics-based one (\emph{i.e.} the shading color computed from the predicted material, normal, the co-located flashlight, and the ambient light).
We empirically find this strategy makes the geometry reconstruction more robust.

\subsection{Modeling Photographer's Occlusion in Combined Light Representation}\label{sec:occ}
When capturing real-world data, we find the photographer would occlude the ambient light when he or she holds the camera moving around the target subject.
In an environment with moderate-level ambient illumination (\emph{e.g. noon w/o curtain} and \emph{asym. ambient}), the photographer's occlusion becomes more apparent.
In this scenario, using only $K_{lm}$ to represent the ambient shading is inadequate, as the ambient light is changed as the photographer moves.
Thus, we propose to explicitly model the photographer.

Inspired by Eclipse~\cite{verbin2023eclipse}, we assign each training view a learnable occlusion mask parameterized as 2-order Spherical Harmonics (SH).
Thus, the ambient shading for the $i$-th view becomes:
\begin{equation}
    l_{amb} = c\cdot O_{amb}^i(\textbf{n})\cdot {\rm SoftPlus}(\sum_{l=0}^2\sum_{m=-l}^{l}\cdot K_{lm}\cdot Y_{lm}(\textbf{n}))
\end{equation}
Here, $O_{amb}^i(\cdot)$ is the visibility mask for the $i$-th view to compensate for the occlusion caused by the photographer, parameterized as:
\begin{equation}
    O_{amb}^i(\textbf{n}) = {\rm Sigmoid}(\sum_{l=0}^2\sum_{m=-l}^{l}\cdot O^i_{lm}\cdot Y_{lm}(\textbf{n}))
\end{equation}
Here, $O^i_{lm}$ are the SH coefficients for the occlusion mask for the $i$-th view, which are learned together with the $K_{lm}$.
In this way, we adopt $K_{lm}$ to represent the global ambient shading and a per-view $O^i_{lm}$ to represent the photograph's occlusion.

\subsection{Losses}
We detail all the loss functions we used in the following.
At a specific training iteration, we cast $n$ camera rays to the 3D scene.
For the $i$-th ray, we sample $k_i$ points along it according to the empty space skip strategy proposed by the Instant-NGP paper~\cite{muller2022instant}.

\paragraph{Photometric Terms.}
We adopt an L1 photometric loss:
\begin{equation}
    \mathcal{L}_{L1} = \sum_{i=1}^n ||\hat{I}_i - I_i||_1
\end{equation}
Here, $\hat{I}_i$ is the rendered color for the $i$-th ray while $I_i$ is the corresponding ground truth.

In addition, we adopt an LPIPS loss $\mathcal{L}_{lpips}$~\cite{zhang2018perceptual} over the full image to reconstruct richer details.

Both photometric terms are computed in the linear space; we apply a gamma function to convert the sRGB space into the linear space and empirically set the gamma value to $2.2$.
Although the LPIPS network is trained on images in the sRGB space, we empirically find it also works well on images in the linear space.

\paragraph{Mask Loss.}
We adopt an L1 mask loss:
\begin{equation}
    \mathcal{L}_{mask} = \sum_{i=1}^n ||\hat{O}_i - O_i||_1
\end{equation}
Here, $\hat{O}_i$ is the rendered occupancy for the $i$-th ray; $O_i$ is the corresponding pseudo ground truth computed from an off-the-shelf face parsing network~\cite{lin2021roi}.

\paragraph{Eikonal Loss.}
We add an Eikonal term~\cite{icml2020_2086} to the sampled points to regularize the SDF value predicted by $f_{sdf}$:
\begin{equation}
    \mathcal{L}_{eikonal} = \sum_{i=1}^n \sum_{j=1}^{k_i} (||\nabla f_{sdf}(\mathbf{x}_{ij})||_2 - 1)^2
\end{equation}

\paragraph{Normal Smooth Loss.}
We add a regularization term to encourage smooth normal by constraining the normal of a sampled point $\mathbf{x}$ to be similar to its nearby point $\mathbf{x}^\epsilon$~\cite{zhang2021nerfactor,rosu2023permutosdf}:
\begin{equation}
    \mathcal{L}_{eps} = \sum_{i=1}^n \sum_{j=1}^{k_i} (1 - \mathbf{n}(\mathbf{x}_{ij}) \cdot \mathbf{n}(\mathbf{x}_{ij}^\epsilon))
\end{equation}
During training, the nearby points are sampled following the strategy of PermutoSDF~\cite{rosu2023permutosdf}.

\paragraph{Composition Loss.}
To constrain the training of our hybrid representation, inspired by ObjectSDF++~\cite{wu2023objectsdf++}, we render the occlusion-aware object opacity mask $\hat{O}^{E}$ and $\hat{O}^{S}$ for the $E$ and $S$ region and compare them to the corresponding ground truth ${O}^{E}$ and ${O}^{S}$ obtained from an off-the-shelf face parsing network~\cite{lin2021roi} over the $n$ sampled rays:
\begin{equation}
    \mathcal{L}_{comp} = \sum_{i=1}^n||\hat{O}^{E}_i - O^{E}_i||_1 + \sum_{i=1}^n||\hat{O}^{S}_i - O^{S}_i||_1
\end{equation}

\begin{figure}[t]
    \centering
    \includegraphics[width=0.475\textwidth]{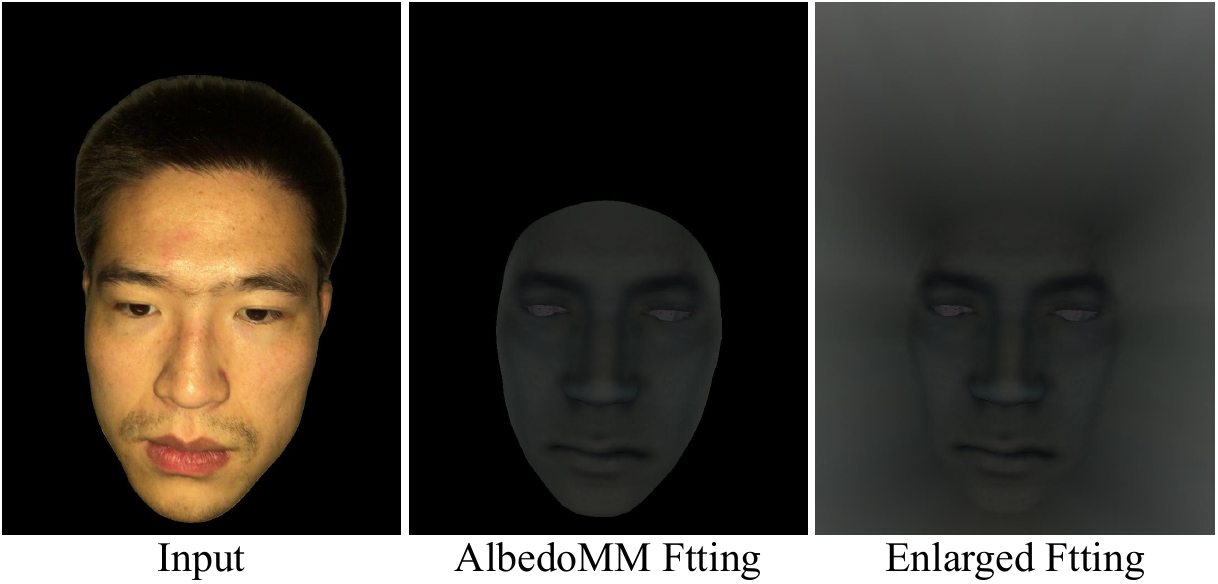}
    \vspace{-15pt}
    \caption{We enlarge the AlbedoMM fitted specular albedo to the whole image as the pseudo ground truth. }
    \label{Fig:ref_loss}
\end{figure}

\paragraph{Reflectance Regularization.}
We exploit the morphable face albedo model -- AlbedoMM~\cite{smith2020morphable} -- as the reflectance prior.
Specifically, we devise a multi-view AlbedoMM fitting algorithm to reconstruct the specular albedo for each frame.
Then, we enlarge the solved specular albedo to the whole image (see Figure~\ref{Fig:ref_loss}) to obtain $I^s$ as pseudo ground truth to supervise the volume-rendered one $\hat{I}^s$ on the sampled rays:
\begin{equation}
    \mathcal{L}_{ref} = \sum_{i=1}^n||k\cdot \hat{I}^s_i - I^s_i||_1
\end{equation}
Here, $k\in\mathbb{R}$ is a learnable scalar to compensate for the scale ambiguity stemming from our predefined light intensity $L$.
For pixels from the eyeballs region $E$, we do not compute $\mathcal{L}_{ref}$ since we already have predefined prior $s_{eye}$. 
For pixels from the hair region indicated by the parsing mask~\cite{lin2021roi}, we constrain its specular albedo to be $0$ to obtain a diffuse appearance as we empirically find fitting a specular lobe produces artifacts when rendered in novel environments.

\subsection{Training Schedule}
We adopt a two-stage training strategy.
In the first stage, we volume render the neural field to optimize the geometry and reflectance network, \emph{i.e.} $f_{sdf}$ and $f_{brdf}$, and the shading coefficients $K_{lm}$ (and the optional $O_{lm}^i$) jointly.
We set $\omega_{L1}$ to $1$, $\omega_{mask}$ to $1$, $\omega_{eikonal}$ to $1$, $\omega_{eps}$ to $0.5$ in the hair region while $0.02$ in other region, $\omega_{comp}$ to $1$, and $\omega_{ref}$ to $0.5$.

In the second stage, we extract the mesh from the neural field and perform surface rendering; we fix the geometry network $f_{sdf}$ while only optimizing the reflectance network $f_{brdf}$ and the shading coefficients $K_{lm}$ (and the optional $O_{lm}^i$) in this stage.
Since geometry is fixed, only photometric loss and reflectance regularization are adopted.
We set $\omega_{L1}$ to $1$, $\omega_{lpips}$ to $0.1$, and $\omega_{ref}$ to $0.01$.
Note that we turn down the weight of the statistical prior, \emph{i.e.} $\omega_{ref}$, in the second stage to encourage the network to recover more person-specific specular details from the observations.

We train our method by $40000$ iterations.
The first $30000$ iterations are for the first stage and the last $10000$ iterations are for the second stage.
We adopt the Adam optimizer with an initial learning rate of $0.001$.
The learning rate is annealed by $0.3$ for every $15000$ iterations.
Our method can be trained within 70 minutes using a single Nvidia RTX 3090 graphics card.
In our experiment, the loss weights are shared for all the captured sequences.

\subsection{Automatic 3D Assets Extraction}
For the $S$ region, we extract the $0.001$ iso-surface of the neural SDF field using the Marching Cubes~\cite{lorensen1998marching} algorithm as we find the geometry is biased in VolSDF; a similar observation can be found in BakedSDF~\cite{yariv2023bakedsdf}.
For the $E$ region, we directly use the sphere meshes as its geometry.

We rasterize the extracted meshes into all the training views and compare the rendered occupancy mask to the face parsing mask.
For the triangle faces unseen from every training view, we delete them from the extracted meshes.
Then, we find all the connection areas on the mesh using an existing tool~\cite{trimesh}. 
We only keep the largest one while deleting all other connection areas.

We adopt Blender's UV Unwrap function to create the UV mapping function for the meshes.
Then, we generate a normal map, diffuse albedo map, specular albedo map, and roughness map accordingly.

\subsection{Relightable Performance Capture}
In this Section, we show how we combine our method with the Reflectance Transfer technique~\cite{peers2007post} to construct a simple and powerful baseline for the challenging problem of relightable facial performance capture in a low-cost setup.

\paragraph{Preliminary of Reflectance Transfer.}
The core idea of Reflectance Transfer is to use optical flow to transfer the lighting effects of a source frame to a target frame.
This way, one can capture a relightable scan for only one facial expression as the source frame.
Then, a new performance sequence of the same person (or a different person who has a similar appearance to the source person) can be relit.

Specifically, the relightable scan for the source frame is the densely sampled light transport function captured by the Light Stage~\cite{debevec2000acquiring}.
The target performance sequence $\{I^i_{src}\}_{i=1}^n$ is captured under a known lighting $L_{src}$.

To obtain the relit target performance sequence $\{I^i_{tgt}\}_{i=1}^n$ under a new lighting $L_{tgt}$, they first render the source relightable scan under $L_{src}$ and $L_{tgt}$ to obtain the corresponding renderings $I^0_{src}$ and $I^0_{tgt}$.
The lighting effects $R^0$ is defined as the ratio image of $I^0_{src}$ and $I^0_{tgt}$:
\begin{equation}
    R^0 = \frac{I^0_{tgt}}{I^0_{src}}
    \label{eq:ratio}
\end{equation}
Then, they warp the lighting effects $R_0$ to a target frame $I^i_{src}$ using the warping function computed from $I^0_{src}$ and $I^i_{src}$ to obtain $R_i$:
\begin{equation}
    R^i = {\rm warp}(R^0)
\end{equation}
Here, ${\rm warp}(\cdot)$ is the optical flow computed from $I^i_{src}$ to $I^0_{src}$; note that the lighting condition of $I^0_{src}$ and $I^i_{src}$ are the same, which is the key to compute reliable optical flow.

By multiplying the warped light effects $R^i$ with the target frame $I^i_{src}$, a relit frame $I^i_{tgt}$ can be obtained:
\begin{equation}
    I^i_{tgt} = R^i \odot I^i_{src}
\end{equation}

See their paper for other details including refining the warping function, filtering the ratio image, aligning the head pose, and the keyframe propagation technique to enhance temporal consistency.
Although the Reflectance Transfer method is not physically based, it works well for a large body of low or mid-frequency illuminations as demonstrated by their paper~\cite{peers2007post}.

\paragraph{Combine Our Method to Reflectance Transfer.} 
Recall that the Reflectance Transfer method requires a relightable scan for the source frame, a facial performance sequence captured under known lighting $L_{src}$, and a target lighting $L_{tgt}$.
Our goal is to construct a low-cost version of the Reflectance Transfer to support relightable facial performance capture in the low-cost setup.
To this end, we modify their method in several aspects.

For its first requirement -- the relightable scan for the source frame, we can directly replace it with our method's results.
Given the target lighting $L_{tgt}$, we can directly render $I^0_{tgt}$.
However, its second requirement, \emph{i.e.} capturing the facial performance sequence under known lighting, is hard to fulfill in the low-cost setup.
Thus, we propose to capture the performance sequence under unknown but low-frequency lighting and solve it using our relightable scan.

Specifically, we render the source frame under the first $2$-order Spherical Harmonics (SH) basis lighting to obtain $\{I^0_{i}\}_{i=0}^8$.
We parameterize the lighting as the linear combination weights $\{c_{i}\}_{i=0}^8$ of these SH bases.
Given a target frame $I^i_{src}$ captured under the unknown lighting $L_{src}$, our goal is to estimate $\{c_{i}\}_{i=0}^8$ to minimize the following photometric loss:
\begin{equation}
    \mathcal{L}_{pho} = ||{\rm warp}(\hat{I}^0_{src}) - I^i_{src}||_1
\end{equation}
Here, ${\rm warp}(\cdot)$ is the optical flow computed from $I^i_{src}$ to $\hat{I}^0_{src}$ using RAFT~\cite{teed2020raft}; $\hat{I}^0_{src}$ is computed as the linear combination of $\{I^0_{i}\}_{i=0}^8$ weighted by $\{c_{i}\}_{i=0}^8$:
\begin{equation}
    \hat{I}^0_{src} = \sum_{i=0}^8 c_i \cdot I^0_{i}
\end{equation}

In our scenario, RAFT can be seen as a fully differentiable optical flow solver. 
Thus, the photometric loss function is fully differentiable \emph{w.r.t} the unknowns, \emph{i.e.} $\{c_{i}\}_{i=0}^8$.
We adopt gradient descent to minimize $\mathcal{L}_{pho}$.
Note that in each iteration step, $\hat{I}^0_{src}$ are updated since $\{c_{i}\}_{i=0}^8$ are updated.
Thus, the warping function also needs to be re-computed in each iteration. 

At the beginning of the optimization, the lighting of $\hat{I}^0_{src}$ may be far from $I^i_{src}$, which violates the optical flow assumption.
However, we empirically find RAFT can also produce reasonable results in this scenario.
During the optimization process, we observe that as the lighting of $\hat{I}^0_{src}$ becomes closer to $I^i_{src}$, the estimated optical flow becomes more accurate.

So far, we have introduced how to compute $I^0_{src}$ and $I^0_{tgt}$. 
Thus, we can define the light effect $R_{0}$ as Equation~\eqref{eq:ratio}.
In this way, we can directly adopt Reflectance Transfer to relight the whole facial performance sequence.

\paragraph{Results.}
See our \emph{supplementary video} for the facial performance relighting results.

\paragraph{Limitations and Discussions.}
Although impressive results are demonstrated, this baseline method for low-cost facial performance capture has several limitations. 

Similar to the original Reflectance Transfer method, all the shading effects, \emph{e.g.} specularities, shadows, are transferred in the image space via optical flow.
Thus, there is no guarantee that these effects are conform to the geometry.
For example, we observe the hard shadow flickered across the frames when relit under a high-frequency target illumination due to the unstable optical flow estimation.
We emphasize that facial performance relighting under high-frequency illuminations is very challenging even has access to the Light Stage data~\cite{pandey2021total,zhang2021video,meka2019deep}.
When relit under low-frequency illuminations, we observe it can often produce plausible results.

Another limitation inherited from the original Reflectance Transfer method is that, as there is only one source frame, in a target frame there may be some regions that can not find correspondence in the source frame.
For example, if eyes in the source frame are opened up, a closed-eye target frame's relighting effects around eyelids cannot be found in the source frame; in this case, the eyelids' lighting effects are hallucinated via the warping function.

Nevertheless, we believe our solution can serve as a strong baseline in the field of low-cost facial performance relighting to motivate future work.

\section{More Experiments}

\subsection{More Evaluations}

\begin{figure*}[t]
    \centering
    \includegraphics[width=\textwidth]{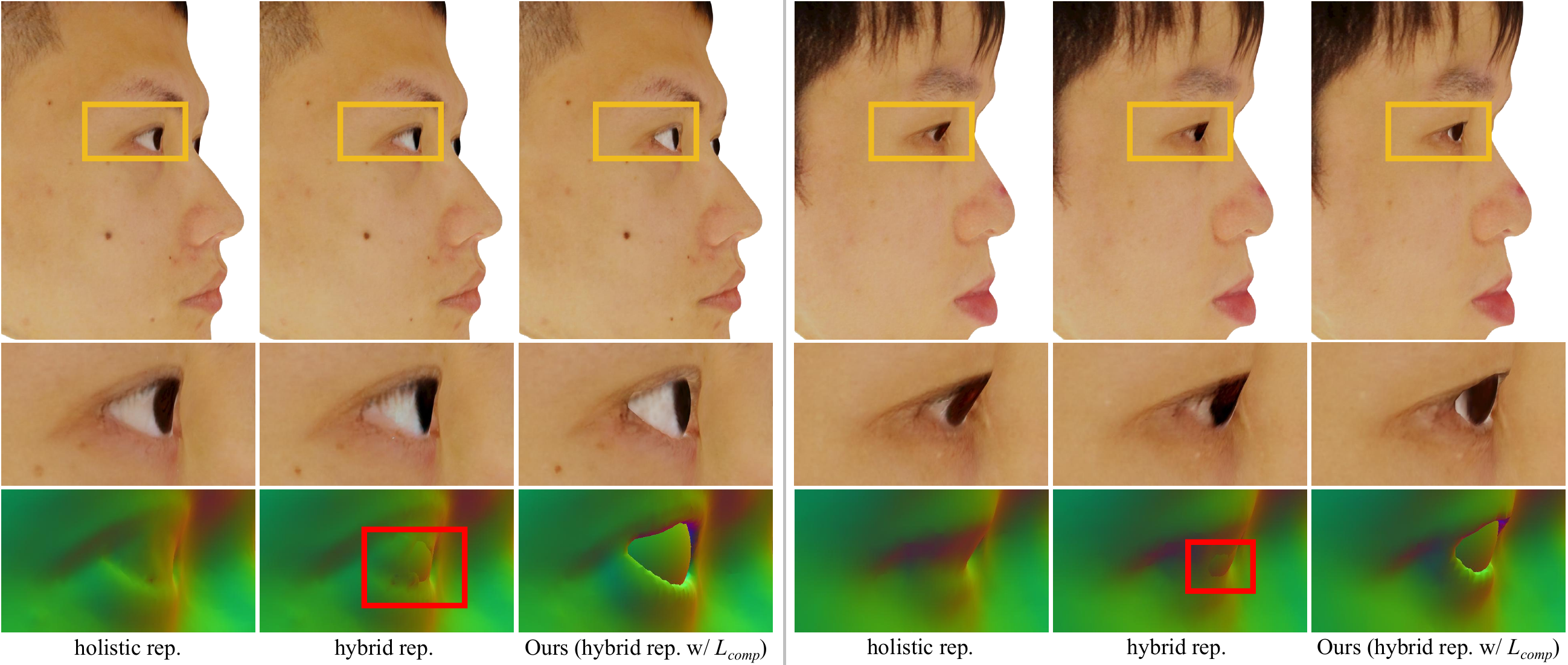}
    \vspace{-15pt}
    \caption{More qualitative evaluation of the hybrid representation and $\mathcal{L}_{comp}$ on geometry reconstruction around eyes. The close-up rendered texture and normal are shown in the second and third rows respectively. }
    \label{Fig:eval_hybrid}
\end{figure*}

\begin{figure*}[t]
    \centering
    \includegraphics[width=\textwidth]{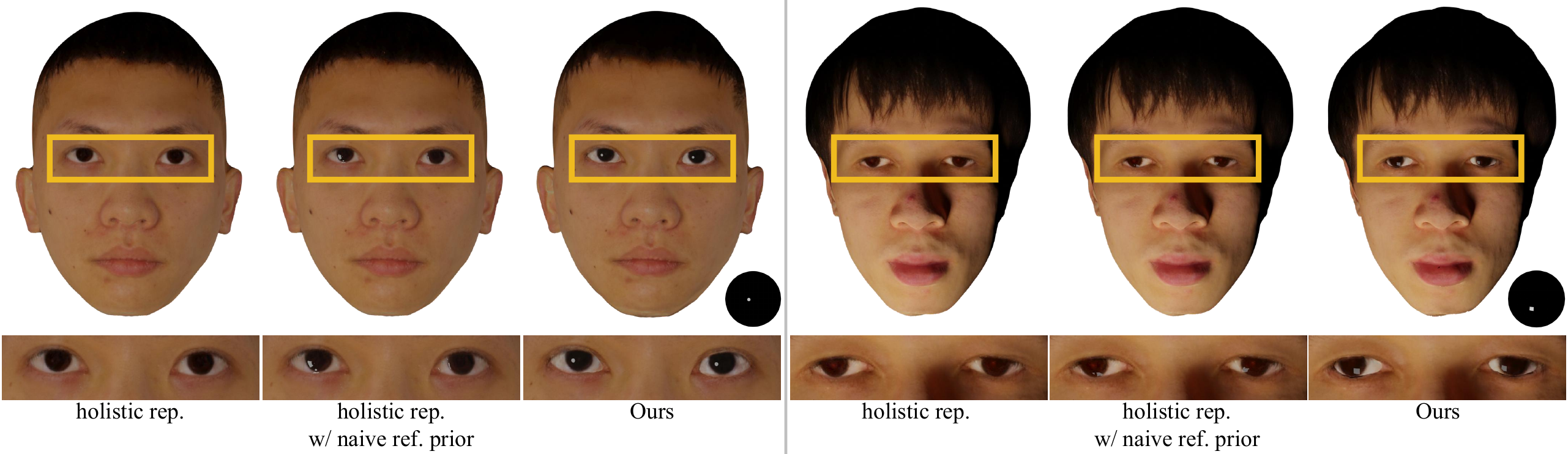}
    \vspace{-15pt}
    \caption{More qualitative evaluation of our hybrid representation and the baseline variants on relighting. }
    \label{Fig:eval_hybrid_relight}
\end{figure*}

\paragraph{Hybrid Face Representation and $\mathcal{L}_{comp}$.}
We show more qualitative evaluations of our hybrid representation and the composition loss $\mathcal{L}_{comp}$ in Figure~\ref{Fig:eval_hybrid} and Figure~\ref{Fig:eval_hybrid_relight}.

\begin{figure}[t]
    \centering
    \includegraphics[width=0.475\textwidth]{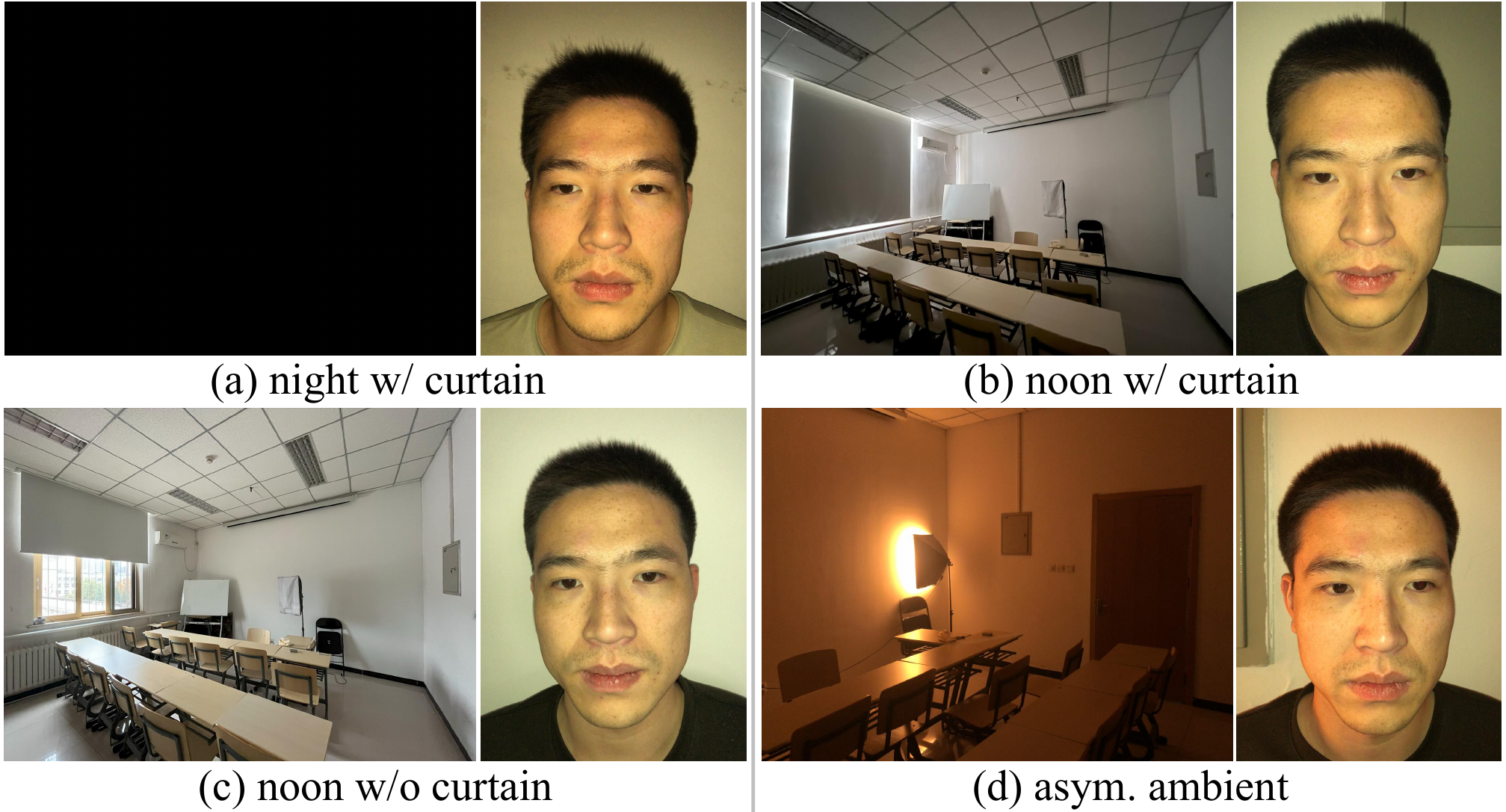}
    \vspace{-15pt}
    \caption{The photo of the scenes we conducted experiments on and the corresponding example images captured in the scene. Note that \emph{asym. ambient} is only referenced in the \emph{supplementary material} while the other three scenes are mentioned in the main paper. }
    \label{Fig:scenes}
\end{figure}

\begin{figure}[t]
    \centering
    \includegraphics[width=0.475\textwidth]{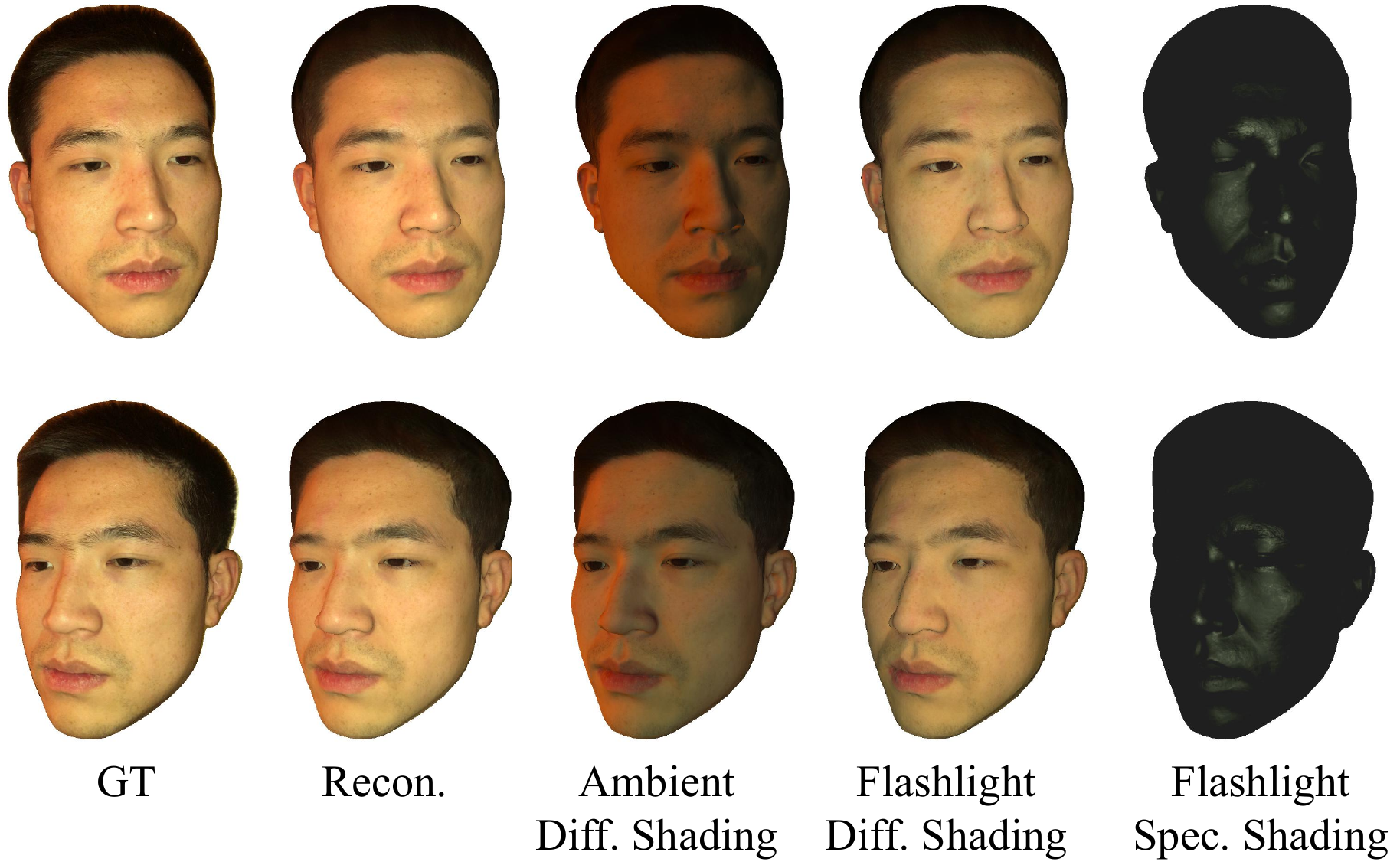}
    \vspace{-15pt}
    \caption{The reconstructed shading contribution of the ambient and the smartphone flashlight on two viewpoints. This experiment is conducted on the data captured in \emph{asym. ambient}.}
    \label{Fig:exp-light_rep_comp}
\end{figure}

\begin{figure}[t]
    \centering
    \includegraphics[width=0.475\textwidth]{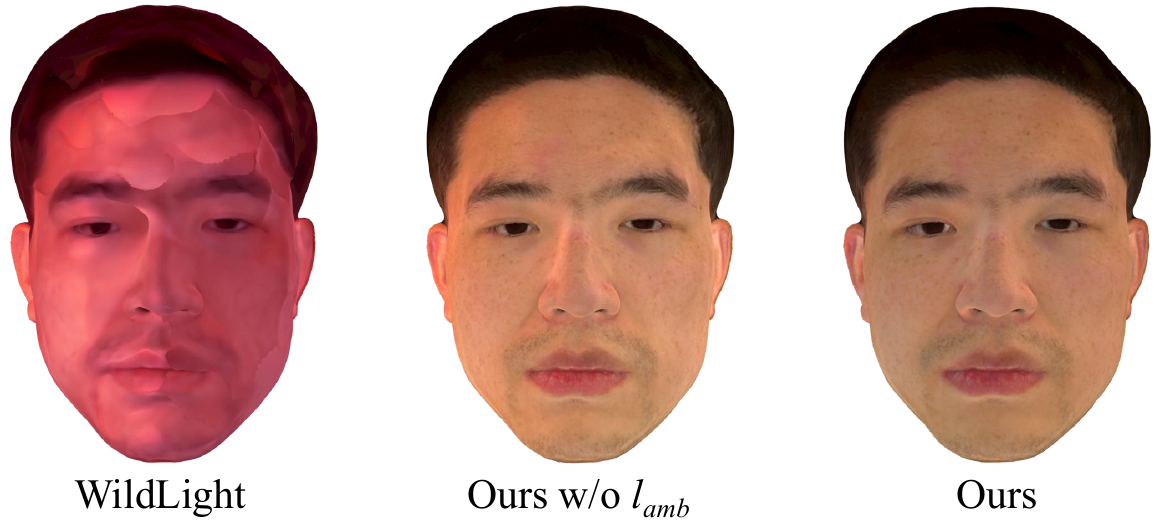}
    \vspace{-15pt}
    \caption{Comparison on diffuse albedo reconstruction of WildLight, our method, and the baseline variant. This experiment is conducted on the data captured in \emph{asym. ambient}. }
    \label{Fig:cmp-light_rep}
\end{figure}

\paragraph{Combined Light Representation.}
In Figure~\ref{Fig:scenes}, we show the photo of the three captured environments mentioned in Section 4.1 (Combined Light Representation) of our main paper, \emph{i.e.} \emph{night w/ curtain}, \emph{noon w/ curtain}, and \emph{noon w/o curtain}.
In the following, we present a more thorough evaluation of this design choice.

We conduct experiments on a challenging real scene with asymmetric ambient illumination; we dub it \emph{asym. ambient}. 
In this real scene, a red lamp is placed on the right side of the face.
The photo of this scene and one sampled captured frame can be found in Figure~\ref{Fig:scenes}.
We explicitly model the occlusion caused by the photographer (see Section~\ref{sec:occ}) when testing on this scene.
We show the reconstructed shading component of the ambient and the smartphone flashlight in Figure~\ref{Fig:exp-light_rep_comp}.
We observe that our method can still disentangle the ambient and flashlight contributions from the captured images in a plausible way.
As shown in Figure~\ref{Fig:cmp-light_rep}, we observe that the baseline variant, \emph{i.e.} \emph{Ours w/o $l_{amb}$}, bakes the red ambient light into the diffuse albedo while our method can obtain a cleaner one.

\begin{figure*}[t]
    \centering
    \includegraphics[width=\textwidth]{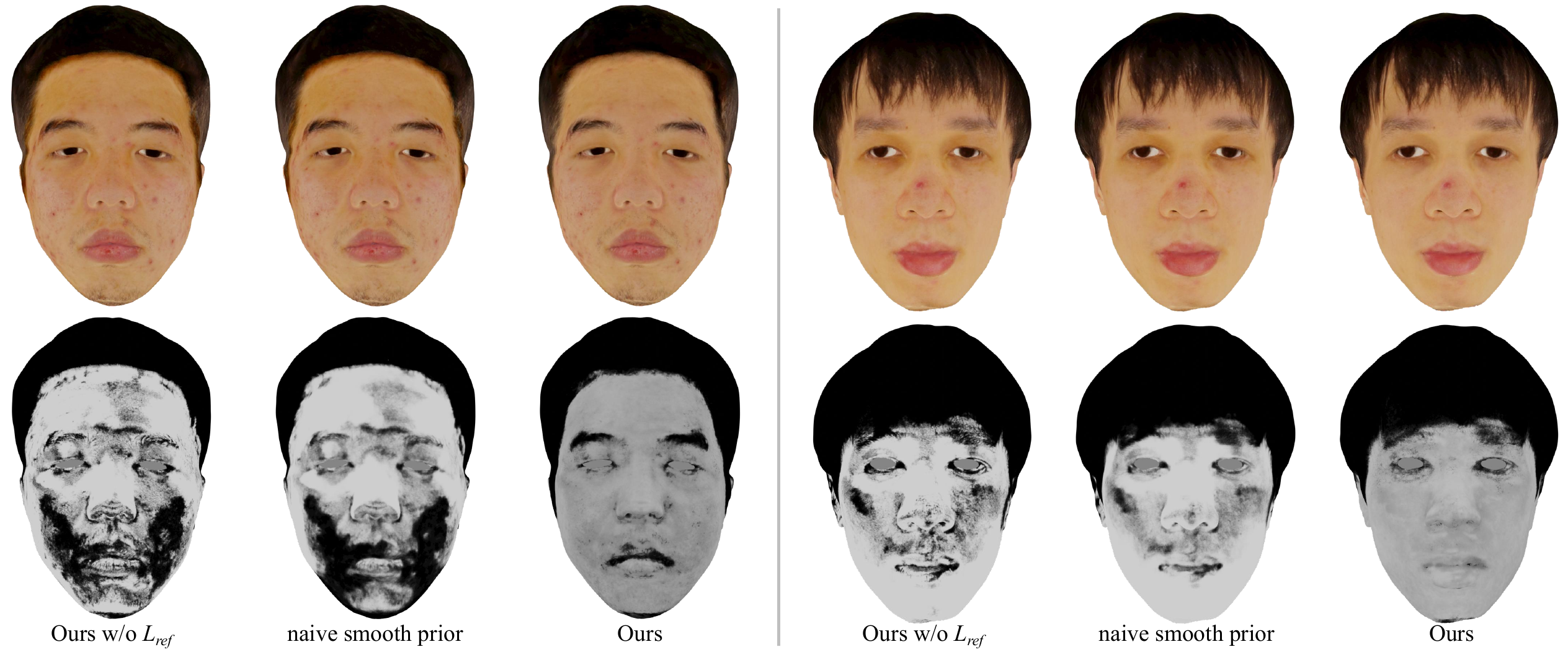}
    \vspace{-15pt}
    \caption{More qualitative evaluation of our reflectance regularization loss $\mathcal{L}_{ref}$ and the baseline variants on diffuse (the first row) and specular (the second row) albedo estimation. }
    \label{Fig:eval_abdmm}
\end{figure*}

\paragraph{Reflectance Regularization.}
We show more qualitative evaluations of our reflectance regularization $\mathcal{L}_{ref}$ in Figure~\ref{Fig:eval_abdmm}.

\begin{table}[t]
\centering
\begin{tabular}{@{}lllll@{}}
\toprule
 & PSNR~$\uparrow$ & SSIM~\cite{wang2004image}~$\uparrow$ & 
 LPIPS~\cite{zhang2018perceptual}~$\downarrow$  \\ \midrule
 NextFace++ & 17.62 & 0.7339 & 0.2727 \\ 
 Wildlight & 23.80 & 0.8205 & 0.2798  \\ \midrule
 Ours & \textbf{26.12} & \textbf{0.8808} & \textbf{0.1642}  \\ \bottomrule
\end{tabular}
\caption{Quantitative comparison of our method and several competitors on face reconstruction. The metric is averaged on $5$ subjects.}
\label{tab:cmp_supp}
\end{table}

\begin{figure}[t]
    \centering
    \includegraphics[width=0.475\textwidth]{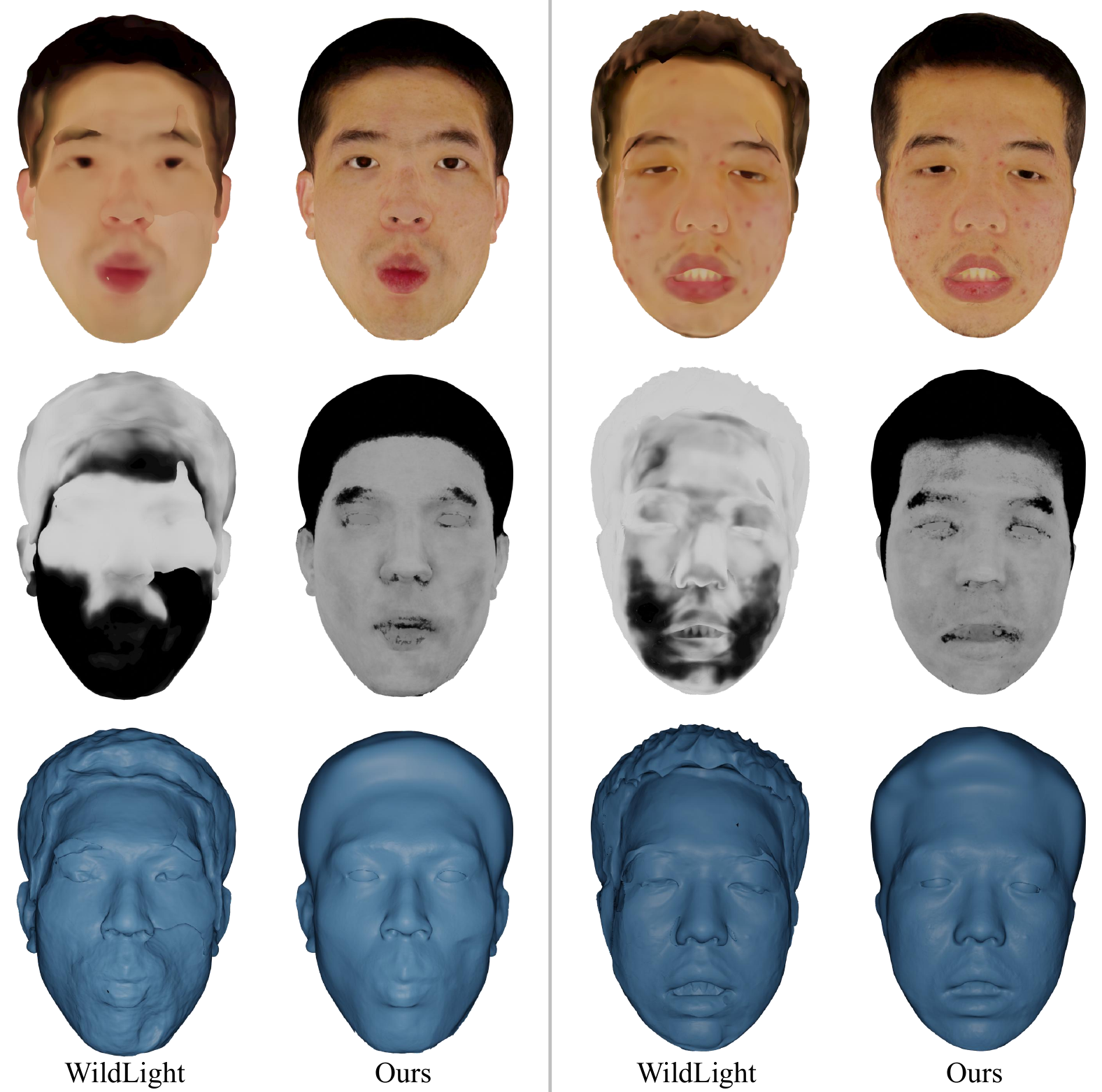}
    \vspace{-15pt}
    \caption{Qualitative comparison of WildLight~\cite{cheng2023wildlight} and our method on diffuse albedo reconstruction (the first row), specular albedo reconstruction (the second row), and geometry reconstruction (the last row). }
    \label{Fig:wl_cmp}
\end{figure}

\subsection{More Comparisons}
\paragraph{Comparison to WildLight.}
We compare our method to WildLight~\cite{cheng2023wildlight}, a state-of-the-art inverse rendering method for generic objects that takes two co-located smartphone flashlight sequences as input, one with the flashlight opened and the other closed.
In WildLight, they learn a NeRF~\cite{mildenhall2020nerf} to model the ambient shading and directly use the flashlight-closed sequence to supervise it.
Compared to WildLight, our method only needs a single flashlight-opened sequence for training as our combined light representation is more compact.
In addition, WildLight cannot model the indirect illumination brought by the smartphone flashlight as their ambient shading NeRF is supervised by the flashlight-closed images.

We first conduct experiments on the data captured in the room at night, \emph{i.e. night w/ curtain}.
As shown in Figure~\ref{Fig:scenes}, in this scenario the images are totally black if the flashlight is closed\footnote{In this case, the task of the ambient shading term $l_{amb}$ in our combined light representation is to model the indirect illumination caused by the smartphone flashlight.}.
Thus, we turn off the ambient shading NeRF in WildLight.
We compare the reconstructed geometry and reflectance in Figure~\ref{Fig:wl_cmp}.
Our method obtains superior results over WildLight since (1) we naturally integrate facial geometry and reflectance priors into our method, leading to better eyeball reconstruction and reflectance estimation, and (2) other design choices of our method, \emph{e.g.} the view-dependent color head and the two-stage training strategy, make it more robust on real-world data and produce more detailed textures.
In Table~\ref{tab:cmp_supp}, we present quantitative results on face reconstruction; to ensure a fair comparison, all the metrics are computed on the mesh renderings.
Not surprisingly, our method obtains better results than WildLight again.
We also copy the NextFace++'s results from the main paper for reference.

We then conduct experiments on the data captured in the challenging \emph{asym. ambient} (see the photo of this scene in Figure~\ref{Fig:scenes}).
We turn on the ambient shading NeRF in WildLight.
We use the same data as our method to train WildLight, \emph{i.e.} a single co-located sequence with the flashlight opened.
As shown in Figure~\ref{Fig:cmp-light_rep}, without direct supervision on the ambient shading NeRF, WildLight cannot disentangle the ambient illumination and the smartphone flashlight from the captured images in a plausible way, leading to an unreasonable diffuse albedo.
Our method obtains better results as our combined light representation is compact enough to disentangle the ambient and flashlight contributions solely from the captured data.

\begin{figure}[t]
    \centering
    \includegraphics[width=0.475\textwidth]{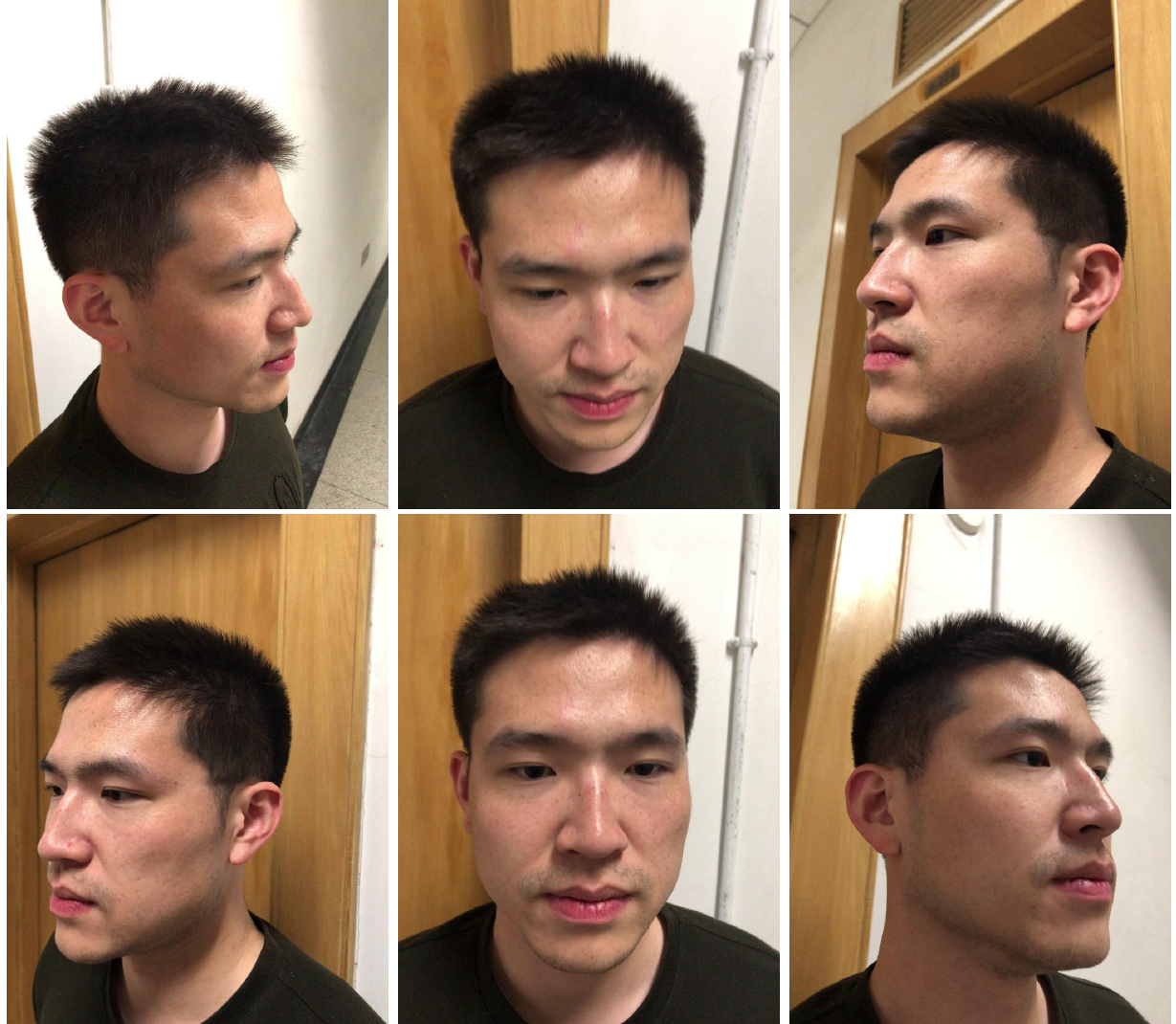}
    \vspace{-15pt}
    \caption{Example multi-view images to train NeRO. These images are captured in an uncontrolled indoor environment. }
    \label{Fig:nero_input}
\end{figure}

\paragraph{Comparison to NeRO}
Some methods~\cite{zhang2021nerfactor,Munkberg_2022_CVPR,hasselgren2022shape,liu2023nero} in inverse rendering take multi-view images of an object as input; from these images, they estimate the environment lighting and the object's geometry and reflectance.
Compared to our method, these works have an even more easy-to-use capture setup for daily users; it neither requires the ambient illumination to be low frequency nor needs the flashlight to be opened up during capture.
Among these works, NeRO~\cite{liu2023nero} is the state-of-the-art.
In this part, we compare our method to NeRO to see whether our setup is necessary to reconstruct high-quality facial geometry and appearance.

\begin{figure}[t]
    \centering
    \includegraphics[width=0.475\textwidth]{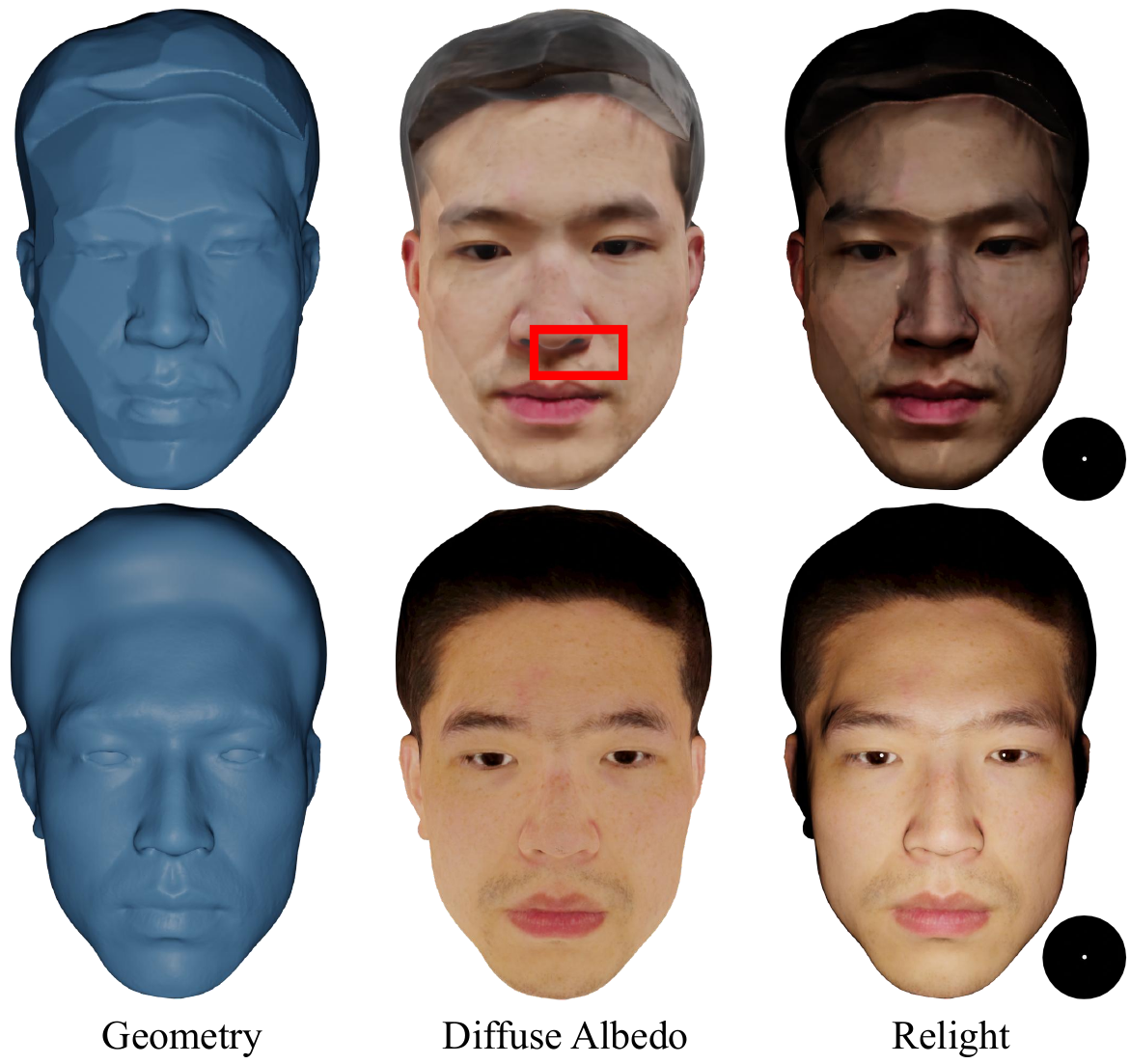}
    \vspace{-15pt}
    \caption{Qualitative comparison of NeRO (the first row) and our method (the second row) on geometry reconstruction, diffuse albedo reconstruction, and relighting results. }
    \label{Fig:nero_cmp}
\end{figure}

We capture two videos for the same identity, one in our capture setup (\emph{i.e.} co-located smartphone flashlight video captured around the subject in a dim room) and the other following NeRO's capture setup (\emph{i.e.} smartphone video captured around the subject in an unconstrained environment).
Some recorded data for training NeRO is shown in Figure~\ref{Fig:nero_input}.
The comparison results are shown in Figure~\ref{Fig:nero_cmp}.

On diffuse albedo reconstruction, NeRO bakes shadow in the diffuse albedo as it is very challenging to simulate the global light transport to model shadow in the inverse rendering process.
Our method obtains a cleaner one as in our capture setup the recorded images are almost shadow-free.
On geometry reconstruction, our method obtains better results again since our setup makes the inverse rendering problem easier compared to NeRO as we have prior knowledge on lighting, \emph{i.e.} the combination of a low-frequency ambient and a high-frequency flashlight.
On face relighting, our method demonstrates better results for two reasons: (1) our method can estimate more accurate geometry and reflectance, and (2) our hybrid face representation can better model the eyes than using a single neural SDF to represent the whole face.

\paragraph{Discussion of Other Related Works}
NeuFace~\cite{zheng2023neuface} is a recent method proposed to reconstruct geometry and neural BRDF from multi-view images of a subject captured in a studio with the synchronized multi-camera system.
We do not compare it as our goal is to reconstruct high-quality relightable 3D face assets compatible with common graphics software under a low-cost and easy-to-use capture setup.
However, NeuFace's neural BRDF representation and the corresponding customized shader cannot achieve our goal.

A concurrent work~\cite{rainer2023neural} proposes a method for facial inverse rendering from smartphone-captured multi-view images captured in arbitrary unknown lighting.
However, similar to PolFace~\cite{azinovic2023high} and SunStage~\cite{wang2023sunstage}, they only focus on facial skin capture, while our method proposes a hybrid face representation that can efficiently represent the complete face with skin, mouth interior, hair, and eyes.

\section{Limitations, Discussions, and Future Works}
Although our method demonstrates high-quality facial geometry and appearance capture results under a low-cost and easy-to-use setup, it still has some limitations.

Similar to EyeNeRF~\cite{li2022eyenerf}, the position and radius of the eyeball meshes are manually set in our method, which incurs some manual effort to the whole pipeline.
We have tried to optimize the eyeballs' position and radius in the training process in our preliminary experiment.
However, we find the results are not always plausible since the gaze direction is the same across the captured frames, which cannot provide enough cues to solve accurate eyeballs.
In future work, we plan to capture a sequence of multi-gaze data before the face capture process to solve the position and radius of the eyeball automatically~\cite{wen2020accurate}.

Although plausible results are obtained, the geometry and reflectance model of the eyeball are still not very accurate.
In the previous work on high-quality eyeball capture~\cite{berard2014high}, a more complex model was designed for eyeballs.
In fact, our hybrid representation makes no assumption on the eyeballs' mesh; to support other eyeball geometry we only need to modify the way to convert the meshes to the SDF field.
We chose the sphere meshes because it is easy to use by daily users while the more complex and accurate one~\cite{berard2014high} is not publicly available.
Replacing the sphere eyeball model with this more complex and accurate one in our method to enhance the realism of eye rendering is an interesting direction.

We observe artifacts in the close-up view of the reconstructed eyeballs and/or other face regions' diffuse albedo.
We attribute this to the following reasons: (1) our data capture setup requires around 25 seconds to record the sequence for a subject, some face parts (\emph{e.g.} the eyelids, the lips, and the tongue) would move inevitably during the capture although we manually remove some frames with apparent movement like eyes blinking, (2) our method relies on off-the-shelf methods to provide segmentation mask and camera calibrations, which are not perfect, and (3) our method does not model eyelashes so they are often baked in the eyeballs' diffuse albedo. 
However, we find such artifacts are often imperceptible in the whole face's scale.

\begin{figure}[t]
    \centering
    \includegraphics[width=0.475\textwidth]{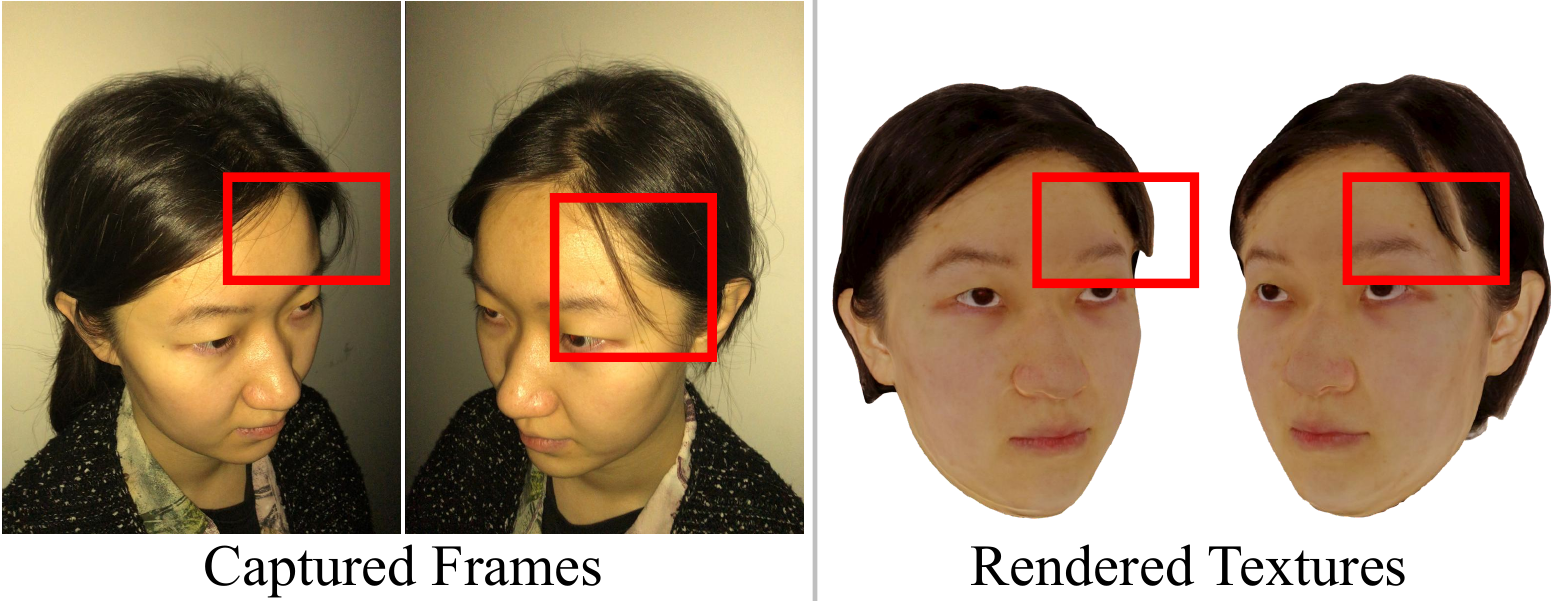}
    \vspace{-15pt}
    \caption{Our method cannot reconstruct plausible flying hairs. }
    \label{Fig:lim_hair}
\end{figure}

We assume the hair is a diffuse surface currently.
We believe using a more accurate reflectance model~\cite{chiang2015practical} to represent hair is a very interesting future work.
However, it is quite challenging in the low-cost setup since we do not have enough cues to reconstruct accurate hair geometry and reflectance~\cite{sun2021human}.
Our method produces artifacts when representing flying hairs as shown in Figure~\ref{Fig:lim_hair}. 
We emphasize that it is a very challenging problem even in studio-based hair reconstruction methods~\cite{sun2021human,nam2019strand}.

{
    \small
    \bibliographystyle{ieeenat_fullname}
    \bibliography{main}

@String(CVPR= {IEEE Conf. Comput. Vis. Pattern Recog.})

@String(ICCV= {Int. Conf. Comput. Vis.})

@String(ECCV= {Eur. Conf. Comput. Vis.})

@String(TOG= {ACM Trans. Graph.})

@String(CVPR  = {CVPR})

@String(ICCV  = {ICCV})

@String(ECCV  = {ECCV})

@String(TOG   = {ACM TOG})

@inproceedings{rosu2023permutosdf,
  title={Permutosdf: Fast multi-view reconstruction with implicit surfaces using permutohedral lattices},
  author={Rosu, Radu Alexandru and Behnke, Sven},
  booktitle={Proceedings of the IEEE/CVF Conference on Computer Vision and Pattern Recognition},
  pages={8466--8475},
  year={2023}
}

@incollection{icml2020_2086,
 author = {Gropp, Amos and Yariv, Lior and Haim, Niv and Atzmon, Matan and Lipman, Yaron},
 booktitle = {Proceedings of Machine Learning and Systems 2020},
 pages = {3569--3579},
 title = {Implicit Geometric Regularization for Learning Shapes},
 year = {2020}
}

@article{lin2021roi,
    title = {RoI Tanh-polar transformer network for face parsing in the wild},
    journal = {Image and Vision Computing},
    volume = {112},
    pages = {104190},
    year = {2021},
    issn = {0262-8856},
    doi = {https://doi.org/10.1016/j.imavis.2021.104190},
    url = {https://www.sciencedirect.com/science/article/pii/S0262885621000950},
    author = {Yiming Lin and Jie Shen and Yujiang Wang and Maja Pantic},
}

@inproceedings{burley2012physically,
  title={Physically-based shading at disney},
  author={Burley, Brent and Studios, Walt Disney Animation},
  booktitle={Acm Siggraph},
  volume={2012},
  pages={1--7},
  year={2012},
  organization={vol. 2012}
}

@inproceedings{cheng2023wildlight,
  title={WildLight: In-the-wild Inverse Rendering with a Flashlight},
  author={Cheng, Ziang and Li, Junxuan and Li, Hongdong},
  booktitle={Proceedings of the IEEE/CVF Conference on Computer Vision and Pattern Recognition},
  pages={4305--4314},
  year={2023}
}

@article{muller2022instant,
  title={Instant neural graphics primitives with a multiresolution hash encoding},
  author={M{\"u}ller, Thomas and Evans, Alex and Schied, Christoph and Keller, Alexander},
  journal={ACM Transactions on Graphics (ToG)},
  volume={41},
  number={4},
  pages={1--15},
  year={2022},
  publisher={ACM New York, NY, USA}
}

@article{zhang2021nerfactor,
  title={Nerfactor: Neural factorization of shape and reflectance under an unknown illumination},
  author={Zhang, Xiuming and Srinivasan, Pratul P and Deng, Boyang and Debevec, Paul and Freeman, William T and Barron, Jonathan T},
  journal={ACM Transactions on Graphics (ToG)},
  volume={40},
  number={6},
  pages={1--18},
  year={2021},
  publisher={ACM New York, NY, USA}
}

@inproceedings{zhang2018perceptual,
  title={The Unreasonable Effectiveness of Deep Features as a Perceptual Metric},
  author={Zhang, Richard and Isola, Phillip and Efros, Alexei A and Shechtman, Eli and Wang, Oliver},
  booktitle={CVPR},
  year={2018}
}

@inproceedings{Munkberg_2022_CVPR,
    author    = {Munkberg, Jacob and Hasselgren, Jon and Shen, Tianchang and Gao, Jun and Chen, Wenzheng 
                    and Evans, Alex and M\"uller, Thomas and Fidler, Sanja},
    title     = "{Extracting Triangular 3D Models, Materials, and Lighting From Images}",
    booktitle = {Proceedings of the IEEE/CVF Conference on Computer Vision and Pattern Recognition (CVPR)},
    month     = {June},
    year      = {2022},
    pages     = {8280-8290}
}

@article{hasselgren2022shape,
  title={Shape, light, and material decomposition from images using Monte Carlo rendering and denoising},
  author={Hasselgren, Jon and Hofmann, Nikolai and Munkberg, Jacob},
  journal={Advances in Neural Information Processing Systems},
  volume={35},
  pages={22856--22869},
  year={2022}
}

@article{liu2023nero,
  title={NeRO: Neural Geometry and BRDF Reconstruction of Reflective Objects from Multiview Images},
  author={Liu, Yuan and Wang, Peng and Lin, Cheng and Long, Xiaoxiao and Wang, Jiepeng and Liu, Lingjie and Komura, Taku and Wang, Wenping},
  journal={arXiv preprint arXiv:2305.17398},
  year={2023}
}

@inproceedings{wen2020accurate,
  title={Accurate Real-time 3D Gaze Tracking Using a Lightweight Eyeball Calibration},
  author={Wen, Quan and Bradley, Derek and Beeler, Thabo and Park, Seonwook and Hilliges, Otmar and Yong, Junhai and Xu, Feng},
  booktitle={Computer Graphics Forum},
  volume={39},
  number={2},
  pages={475--485},
  year={2020},
  organization={Wiley Online Library}
}

@article{berard2014high,
  title={High-quality capture of eyes.},
  author={B{\'e}rard, Pascal and Bradley, Derek and Nitti, Maurizio and Beeler, Thabo and Gross, Markus H},
  journal={ACM Trans. Graph.},
  volume={33},
  number={6},
  pages={223--1},
  year={2014}
}

@inproceedings{debevec2000acquiring,
  title={Acquiring the reflectance field of a human face},
  author={Debevec, Paul and Hawkins, Tim and Tchou, Chris and Duiker, Haarm-Pieter and Sarokin, Westley and Sagar, Mark},
  booktitle={Proceedings of the 27th annual conference on Computer graphics and interactive techniques},
  pages={145--156},
  year={2000}
}

@incollection{chiang2015practical,
  title={A practical and controllable hair and fur model for production path tracing},
  author={Chiang, Matt Jen-Yuan and Bitterli, Benedikt and Tappan, Chuck and Burley, Brent},
  booktitle={ACM SIGGRAPH 2015 Talks},
  pages={1--1},
  year={2015}
}

@article{yariv2023bakedsdf,
  title={BakedSDF: Meshing Neural SDFs for Real-Time View Synthesis},
  author={Yariv, Lior and Hedman, Peter and Reiser, Christian and Verbin, Dor and Srinivasan, Pratul P and Szeliski, Richard and Barron, Jonathan T and Mildenhall, Ben},
  journal={arXiv preprint arXiv:2302.14859},
  year={2023}
}

@incollection{lorensen1998marching,
  title={Marching cubes: A high resolution 3D surface construction algorithm},
  author={Lorensen, William E and Cline, Harvey E},
  booktitle={Seminal graphics: pioneering efforts that shaped the field},
  pages={347--353},
  year={1998}
}

@inproceedings{wu2023objectsdf++,
  title={ObjectSDF++: Improved Object-Compositional Neural Implicit Surfaces},
  author={Wu, Qianyi and Wang, Kaisiyuan and Li, Kejie and Zheng, Jianmin and Cai, Jianfei},
  booktitle={Proceedings of the IEEE/CVF International Conference on Computer Vision},
  pages={21764--21774},
  year={2023}
}

@inproceedings{smith2020morphable,
  title={A morphable face albedo model},
  author={Smith, William AP and Seck, Alassane and Dee, Hannah and Tiddeman, Bernard and Tenenbaum, Joshua B and Egger, Bernhard},
  booktitle={Proceedings of the IEEE/CVF Conference on Computer Vision and Pattern Recognition},
  pages={5011--5020},
  year={2020}
}

@article{peers2007post,
  title={Post-production facial performance relighting using reflectance transfer},
  author={Peers, Pieter and Tamura, Naoki and Matusik, Wojciech and Debevec, Paul},
  journal={ACM Transactions on Graphics (TOG)},
  volume={26},
  number={3},
  pages={52--es},
  year={2007},
  publisher={ACM New York, NY, USA}
}

@inproceedings{teed2020raft,
  title={Raft: Recurrent all-pairs field transforms for optical flow},
  author={Teed, Zachary and Deng, Jia},
  booktitle={Computer Vision--ECCV 2020: 16th European Conference, Glasgow, UK, August 23--28, 2020, Proceedings, Part II 16},
  pages={402--419},
  year={2020},
  organization={Springer}
}

@software{trimesh,
	author = {{Dawson-Haggerty et al.}},
	title = {trimesh},
	url = {https://trimsh.org/},
	version = {3.2.0},
	date = {2019-12-8},
}

@incollection{alexander2009digital,
  title={The digital emily project: photoreal facial modeling and animation},
  author={Alexander, Oleg and Rogers, Mike and Lambeth, William and Chiang, Matt and Debevec, Paul},
  booktitle={Acm siggraph 2009 courses},
  pages={1--15},
  year={2009}
}

@incollection{alexander2013digital,
  title={Digital ira: Creating a real-time photoreal digital actor},
  author={Alexander, Oleg and Fyffe, Graham and Busch, Jay and Yu, Xueming and Ichikari, Ryosuke and Jones, Andrew and Debevec, Paul and Jimenez, Jorge and Danvoye, Etienne and Antionazzi, Bernardo and others},
  booktitle={ACM SIGGRAPH 2013 Posters},
  pages={1--1},
  year={2013}
}

@article{ghosh2011multiview,
  title={Multiview face capture using polarized spherical gradient illumination},
  author={Ghosh, Abhijeet and Fyffe, Graham and Tunwattanapong, Borom and Busch, Jay and Yu, Xueming and Debevec, Paul},
  journal={ACM Transactions on Graphics (TOG)},
  volume={30},
  number={6},
  pages={1--10},
  year={2011},
  publisher={ACM New York, NY, USA}
}

@article{ma2007rapid,
  title={Rapid Acquisition of Specular and Diffuse Normal Maps from Polarized Spherical Gradient Illumination.},
  author={Ma, Wan-Chun and Hawkins, Tim and Peers, Pieter and Chabert, Charles-Felix and Weiss, Malte and Debevec, Paul E and others},
  journal={Rendering Techniques},
  volume={2007},
  number={9},
  pages={10},
  year={2007}
}

@article{weyrich2006analysis,
  title={Analysis of human faces using a measurement-based skin reflectance model},
  author={Weyrich, Tim and Matusik, Wojciech and Pfister, Hanspeter and Bickel, Bernd and Donner, Craig and Tu, Chien and McAndless, Janet and Lee, Jinho and Ngan, Addy and Jensen, Henrik Wann and others},
  journal={ACM Transactions on Graphics (ToG)},
  volume={25},
  number={3},
  pages={1013--1024},
  year={2006},
  publisher={ACM New York, NY, USA}
}

@article{riviere2020single,
  title={Single-shot high-quality facial geometry and skin appearance capture},
  author={Riviere, J{\'e}r{\'e}my and Gotardo, Paulo and Bradley, Derek and Ghosh, Abhijeet and Beeler, Thabo},
  year={2020},
  publisher={Association for Computing Machinery (ACM)}
}

@article{gotardo2018practical,
  title={Practical dynamic facial appearance modeling and acquisition},
  author={Gotardo, Paulo and Riviere, J{\'e}r{\'e}my and Bradley, Derek and Ghosh, Abhijeet and Beeler, Thabo},
  year={2018},
  publisher={Association for Computing Machinery}
}

@article{debevec2012light,
  title={The light stages and their applications to photoreal digital actors},
  author={Debevec, Paul},
  journal={SIGGRAPH Asia},
  volume={2},
  number={4},
  pages={1--6},
  year={2012}
}

@inproceedings{wang2023sunstage,
  title={Sunstage: Portrait reconstruction and relighting using the sun as a light stage},
  author={Wang, Yifan and Holynski, Aleksander and Zhang, Xiuming and Zhang, Xuaner},
  booktitle={Proceedings of the IEEE/CVF Conference on Computer Vision and Pattern Recognition},
  pages={20792--20802},
  year={2023}
}

@inproceedings{azinovic2023high,
  title={High-res facial appearance capture from polarized smartphone images},
  author={Azinovi{\'c}, Dejan and Maury, Olivier and Hery, Christophe and Nie{\ss}ner, Matthias and Thies, Justus},
  booktitle={Proceedings of the IEEE/CVF Conference on Computer Vision and Pattern Recognition},
  pages={16836--16846},
  year={2023}
}

@inproceedings{ramamoorthi2001signal,
  title={A signal-processing framework for inverse rendering},
  author={Ramamoorthi, Ravi and Hanrahan, Pat},
  booktitle={Proceedings of the 28th annual conference on Computer graphics and interactive techniques},
  pages={117--128},
  year={2001}
}

@article{wang2021neus,
      title={NeuS: Learning Neural Implicit Surfaces by Volume Rendering for Multi-view Reconstruction}, 
      author={Peng Wang and Lingjie Liu and Yuan Liu and Christian Theobalt and Taku Komura and Wenping Wang},
	  journal={NeurIPS},
      year={2021}
}

@inproceedings{verbin2022ref,
  title={Ref-nerf: Structured view-dependent appearance for neural radiance fields},
  author={Verbin, Dor and Hedman, Peter and Mildenhall, Ben and Zickler, Todd and Barron, Jonathan T and Srinivasan, Pratul P},
  booktitle={2022 IEEE/CVF Conference on Computer Vision and Pattern Recognition (CVPR)},
  pages={5481--5490},
  year={2022},
  organization={IEEE}
}

@inproceedings{xie2022neural,
  title={Neural fields in visual computing and beyond},
  author={Xie, Yiheng and Takikawa, Towaki and Saito, Shunsuke and Litany, Or and Yan, Shiqin and Khan, Numair and Tombari, Federico and Tompkin, James and Sitzmann, Vincent and Sridhar, Srinath},
  booktitle={Computer Graphics Forum},
  volume={41},
  number={2},
  pages={641--676},
  year={2022},
  organization={Wiley Online Library}
}

@inproceedings{zhang2022iron,
  title={Iron: Inverse rendering by optimizing neural sdfs and materials from photometric images},
  author={Zhang, Kai and Luan, Fujun and Li, Zhengqi and Snavely, Noah},
  booktitle={Proceedings of the IEEE/CVF Conference on Computer Vision and Pattern Recognition},
  pages={5565--5574},
  year={2022}
}

@inproceedings{barron2021mip,
  title={Mip-nerf: A multiscale representation for anti-aliasing neural radiance fields},
  author={Barron, Jonathan T and Mildenhall, Ben and Tancik, Matthew and Hedman, Peter and Martin-Brualla, Ricardo and Srinivasan, Pratul P},
  booktitle={Proceedings of the IEEE/CVF International Conference on Computer Vision},
  pages={5855--5864},
  year={2021}
}

@inproceedings{mildenhall2020nerf,
 title={NeRF: Representing Scenes as Neural Radiance Fields for View Synthesis},
 author={Ben Mildenhall and Pratul P. Srinivasan and Matthew Tancik and Jonathan T. Barron and Ravi Ramamoorthi and Ren Ng},
 year={2020},
 booktitle={ECCV},
}

@article{yariv2021volume,
  title={Volume rendering of neural implicit surfaces},
  author={Yariv, Lior and Gu, Jiatao and Kasten, Yoni and Lipman, Yaron},
  journal={Advances in Neural Information Processing Systems},
  volume={34},
  pages={4805--4815},
  year={2021}
}

@inproceedings{oechsle2021unisurf,
  title={Unisurf: Unifying neural implicit surfaces and radiance fields for multi-view reconstruction},
  author={Oechsle, Michael and Peng, Songyou and Geiger, Andreas},
  booktitle={Proceedings of the IEEE/CVF International Conference on Computer Vision},
  pages={5589--5599},
  year={2021}
}

@article{yariv2020multiview,
  title={Multiview neural surface reconstruction by disentangling geometry and appearance},
  author={Yariv, Lior and Kasten, Yoni and Moran, Dror and Galun, Meirav and Atzmon, Matan and Ronen, Basri and Lipman, Yaron},
  journal={Advances in Neural Information Processing Systems},
  volume={33},
  pages={2492--2502},
  year={2020}
}

@inproceedings{barron2022mip,
  title={Mip-nerf 360: Unbounded anti-aliased neural radiance fields},
  author={Barron, Jonathan T and Mildenhall, Ben and Verbin, Dor and Srinivasan, Pratul P and Hedman, Peter},
  booktitle={Proceedings of the IEEE/CVF Conference on Computer Vision and Pattern Recognition},
  pages={5470--5479},
  year={2022}
}

@article{bi2020neural,
  title={Neural reflectance fields for appearance acquisition},
  author={Bi, Sai and Xu, Zexiang and Srinivasan, Pratul and Mildenhall, Ben and Sunkavalli, Kalyan and Ha{\v{s}}an, Milo{\v{s}} and Hold-Geoffroy, Yannick and Kriegman, David and Ramamoorthi, Ravi},
  journal={arXiv preprint arXiv:2008.03824},
  year={2020}
}

@inproceedings{ling2023shadowneus,
  title={Shadowneus: Neural sdf reconstruction by shadow ray supervision},
  author={Ling, Jingwang and Wang, Zhibo and Xu, Feng},
  booktitle={Proceedings of the IEEE/CVF Conference on Computer Vision and Pattern Recognition},
  pages={175--185},
  year={2023}
}

@inproceedings{zhang2021physg,
  title={Physg: Inverse rendering with spherical gaussians for physics-based material editing and relighting},
  author={Zhang, Kai and Luan, Fujun and Wang, Qianqian and Bala, Kavita and Snavely, Noah},
  booktitle={Proceedings of the IEEE/CVF Conference on Computer Vision and Pattern Recognition},
  pages={5453--5462},
  year={2021}
}

@inproceedings{boss2021nerd,
  title={Nerd: Neural reflectance decomposition from image collections},
  author={Boss, Mark and Braun, Raphael and Jampani, Varun and Barron, Jonathan T and Liu, Ce and Lensch, Hendrik},
  booktitle={Proceedings of the IEEE/CVF International Conference on Computer Vision},
  pages={12684--12694},
  year={2021}
}

@article{barron2023zipnerf,
    title={Zip-NeRF: Anti-Aliased Grid-Based Neural Radiance Fields},
    author={Jonathan T. Barron and Ben Mildenhall and 
            Dor Verbin and Pratul P. Srinivasan and Peter Hedman},
    journal={ICCV},
    year={2023}
}

@incollection{beeler2010high,
  title={High-quality single-shot capture of facial geometry},
  author={Beeler, Thabo and Bickel, Bernd and Beardsley, Paul and Sumner, Bob and Gross, Markus},
  booktitle={ACM SIGGRAPH 2010 papers},
  pages={1--9},
  year={2010}
}

@inproceedings{lattas2020avatarme,
  title={AvatarMe: Realistically Renderable 3D Facial Reconstruction" in-the-wild"},
  author={Lattas, Alexandros and Moschoglou, Stylianos and Gecer, Baris and Ploumpis, Stylianos and Triantafyllou, Vasileios and Ghosh, Abhijeet and Zafeiriou, Stefanos},
  booktitle={Proceedings of the IEEE/CVF conference on computer vision and pattern recognition},
  pages={760--769},
  year={2020}
}

@inproceedings{han2023learning,
  title={Learning a 3D Morphable Face Reflectance Model From Low-Cost Data},
  author={Han, Yuxuan and Wang, Zhibo and Xu, Feng},
  booktitle={Proceedings of the IEEE/CVF Conference on Computer Vision and Pattern Recognition},
  pages={8598--8608},
  year={2023}
}

@article{yamaguchi2018high,
  title={High-fidelity facial reflectance and geometry inference from an unconstrained image},
  author={Yamaguchi, Shugo and Saito, Shunsuke and Nagano, Koki and Zhao, Yajie and Chen, Weikai and Olszewski, Kyle and Morishima, Shigeo and Li, Hao},
  journal={ACM Transactions on Graphics (TOG)},
  volume={37},
  number={4},
  pages={1--14},
  year={2018},
  publisher={ACM New York, NY, USA}
}

@inproceedings{zheng2023neuface,
  title={NeuFace: Realistic 3D Neural Face Rendering from Multi-view Images},
  author={Zheng, Mingwu and Zhang, Haiyu and Yang, Hongyu and Huang, Di},
  booktitle={Proceedings of the IEEE/CVF Conference on Computer Vision and Pattern Recognition},
  pages={16868--16877},
  year={2023}
}

@inproceedings{lattas2023fitme,
  title={FitMe: Deep Photorealistic 3D Morphable Model Avatars},
  author={Lattas, Alexandros and Moschoglou, Stylianos and Ploumpis, Stylianos and Gecer, Baris and Deng, Jiankang and Zafeiriou, Stefanos},
  booktitle={Proceedings of the IEEE/CVF Conference on Computer Vision and Pattern Recognition},
  pages={8629--8640},
  year={2023}
}

@inproceedings{li2020learning,
  title={Learning formation of physically-based face attributes},
  author={Li, Ruilong and Bladin, Karl and Zhao, Yajie and Chinara, Chinmay and Ingraham, Owen and Xiang, Pengda and Ren, Xinglei and Prasad, Pratusha and Kishore, Bipin and Xing, Jun and others},
  booktitle={Proceedings of the IEEE/CVF conference on computer vision and pattern recognition},
  pages={3410--3419},
  year={2020}
}

@article{li2022eyenerf,
  title={EyeNeRF: a hybrid representation for photorealistic synthesis, animation and relighting of human eyes},
  author={Li, Gengyan and Meka, Abhimitra and Mueller, Franziska and Buehler, Marcel C and Hilliges, Otmar and Beeler, Thabo},
  journal={ACM Transactions on Graphics (TOG)},
  volume={41},
  number={4},
  pages={1--16},
  year={2022},
  publisher={ACM New York, NY, USA}
}

@article{zheng2023avatarrex,
    title={AvatarRex: Real-time Expressive Full-body Avatars},
    author={Zheng, Zerong and Zhao, Xiaochen and Zhang, Hongwen and Liu, Boning and Liu, Yebin},
    journal={ACM Transactions on Graphics (TOG)},
    volume={42},
    number={4},
    articleno={},
    year={2023},
    publisher={ACM New York, NY, USA}
}

@inproceedings{feng2022capturing,
  title={Capturing and animation of body and clothing from monocular video},
  author={Feng, Yao and Yang, Jinlong and Pollefeys, Marc and Black, Michael J and Bolkart, Timo},
  booktitle={SIGGRAPH Asia 2022 Conference Papers},
  pages={1--9},
  year={2022}
}

@inproceedings{dib2021practical,
  title={Practical face reconstruction via differentiable ray tracing},
  author={Dib, Abdallah and Bharaj, Gaurav and Ahn, Junghyun and Th{\'e}bault, C{\'e}dric and Gosselin, Philippe and Romeo, Marco and Chevallier, Louis},
  booktitle={Computer Graphics Forum},
  volume={40},
  number={2},
  pages={153--164},
  year={2021},
  organization={Wiley Online Library}
}

@InProceedings{Paraperas_2023_ICCV,
    author={Paraperas Papantoniou, Foivos and Lattas, Alexandros and Moschoglou, Stylianos and Zafeiriou, Stefanos},
    title={Relightify: Relightable 3D Faces from a Single Image via Diffusion Models},
    booktitle={Proceedings of the IEEE/CVF International Conference on Computer Vision (ICCV)},
    year={2023}
}

@article{li2022nerfacc,
  title={Nerfacc: A general nerf acceleration toolbox},
  author={Li, Ruilong and Tancik, Matthew and Kanazawa, Angjoo},
  journal={arXiv preprint arXiv:2210.04847},
  year={2022}
}

@inproceedings{gerig2018morphable,
  title={Morphable face models-an open framework},
  author={Gerig, Thomas and Morel-Forster, Andreas and Blumer, Clemens and Egger, Bernhard and Luthi, Marcel and Sch{\"o}nborn, Sandro and Vetter, Thomas},
  booktitle={2018 13th IEEE International Conference on Automatic Face \& Gesture Recognition (FG 2018)},
  pages={75--82},
  year={2018},
  organization={IEEE}
}

@article{wang2004image,
  title={Image quality assessment: from error visibility to structural similarity},
  author={Wang, Zhou and Bovik, Alan C and Sheikh, Hamid R and Simoncelli, Eero P},
  journal={IEEE transactions on image processing},
  volume={13},
  number={4},
  pages={600--612},
  year={2004},
  publisher={IEEE}
}

@inproceedings{stratou2011effect,
  title={Effect of illumination on automatic expression recognition: a novel 3D relightable facial database},
  author={Stratou, Giota and Ghosh, Abhijeet and Debevec, Paul and Morency, Louis-Philippe},
  booktitle={2011 IEEE International Conference on Automatic Face \& Gesture Recognition (FG)},
  pages={611--618},
  year={2011},
  organization={IEEE}
}

@article{sun2021human,
  title={Human hair inverse rendering using multi-view photometric data},
  author={Sun, Tiancheng and Nam, Giljoo and Aliaga, Carlos and Hery, Christophe and Ramamoorthi, Ravi},
  year={2021},
  publisher={The Eurographics Association}
}

@inproceedings{nam2019strand,
  title={Strand-accurate multi-view hair capture},
  author={Nam, Giljoo and Wu, Chenglei and Kim, Min H and Sheikh, Yaser},
  booktitle={Proceedings of the IEEE/CVF Conference on Computer Vision and Pattern Recognition},
  pages={155--164},
  year={2019}
}

@article{pandey2021total,
  title={Total relighting: learning to relight portraits for background replacement},
  author={Pandey, Rohit and Escolano, Sergio Orts and Legendre, Chloe and Haene, Christian and Bouaziz, Sofien and Rhemann, Christoph and Debevec, Paul and Fanello, Sean},
  journal={ACM Transactions on Graphics (TOG)},
  volume={40},
  number={4},
  pages={1--21},
  year={2021},
  publisher={ACM New York, NY, USA}
}

@inproceedings{zhang2021video,
  title={Neural video portrait relighting in real-time via consistency modeling},
  author={Zhang, Longwen and Zhang, Qixuan and Wu, Minye and Yu, Jingyi and Xu, Lan},
  booktitle={Proceedings of the IEEE/CVF International Conference on Computer Vision},
  pages={802--812},
  year={2021}
}

@article{meka2019deep,
  title={Deep reflectance fields: high-quality facial reflectance field inference from color gradient illumination},
  author={Meka, Abhimitra and Haene, Christian and Pandey, Rohit and Zollh{\"o}fer, Michael and Fanello, Sean and Fyffe, Graham and Kowdle, Adarsh and Yu, Xueming and Busch, Jay and Dourgarian, Jason and others},
  journal={ACM Transactions on Graphics (TOG)},
  volume={38},
  number={4},
  pages={1--12},
  year={2019},
  publisher={ACM New York, NY, USA}
}

@article{verbin2023eclipse,
  title={Eclipse: Disambiguating Illumination and Materials using Unintended Shadows},
  author={Verbin, Dor and Mildenhall, Ben and Hedman, Peter and Barron, Jonathan T and Zickler, Todd and Srinivasan, Pratul P},
  journal={arXiv preprint arXiv:2305.16321},
  year={2023}
}

@inproceedings{rainer2023neural,
  title={Neural Shading Fields for Efficient Facial Inverse Rendering},
  author={Rainer, Gilles and Bridgeman, Lewis and Ghosh, Abhijeet},
  booktitle={Computer Graphics Forum},
  pages={e14943},
  year={2023},
  organization={Wiley Online Library}
}

@inproceedings{ramamoorthi2001efficient,
  title={An efficient representation for irradiance environment maps},
  author={Ramamoorthi, Ravi and Hanrahan, Pat},
  booktitle={Proceedings of the 28th annual conference on Computer graphics and interactive techniques},
  pages={497--500},
  year={2001}
}
}

\end{document}